\documentclass[acmsmall,screen]{acmart}

\usepackage[utf8]{inputenc}
\usepackage[T1]{fontenc}
\usepackage[scaled=0.8]{beramono}
\usepackage{amsmath}
\usepackage{amssymb}
\usepackage{booktabs}
\usepackage{xcolor,colortbl}
\usepackage{url}
\usepackage{listings}
\usepackage{paralist}
\usepackage{subcaption}
\usepackage{wrapfig}
\usepackage{enumitem}
\usepackage{multicol}
\usepackage{flushend}
\usepackage{bcprules}
\usepackage{nicefrac}
\usepackage{xfrac}
\usepackage{wrapfig}
\usepackage[normalem]{ulem}

\definecolor{ckeyword}{HTML}{7F0055}
\definecolor{ccomment}{HTML}{3F7F5F}
\definecolor{cstring}{HTML}{2A0099}

\definecolor{aliceblue}{rgb}{0.86, 0.90, 0.95}

\newcommand{\cbox}[1]{\colorbox{aliceblue}{#1}}

\lstdefinelanguage{Scala}%
{morekeywords={abstract,%
  case,catch,char,class,%
  def,else,extends,final,finally,for,%
  if,import,implicit,%
  match,module,lazy,%
  new,null,undefined,%
  override,%
  package,private,protected,public,%
  for,public,return,super,%
  this,throw,trait,try,type,%
  val,var,%
  with,while,%
  let,skip,assert,then,fst,snd,root,idx,sum,prod,exists,forall,%
  yield%
  },%
  sensitive,%
  escapechar=^,
  morecomment=[l]//,%
  morecomment=[s]{/*}{*/},%
  morestring=[b]",%
  showstringspaces=false%
}[keywords,comments,strings]%

\definecolor{listingbg}{RGB}{240, 240, 240}

\newcommand{\commentstyle}[1]{\color{ccomment}\itshape{#1}}
\newcommand{\keywordstyle}[1]{\color{ckeyword}\bfseries{#1}}
\newcommand{\stringstyle}[1]{\color{cstring}\bfseries{#1}}

\lstnewenvironment{listing}{\lstset{language=Scala}}{}
\lstnewenvironment{listingtiny}{\lstset{language=Scala,basicstyle=\scriptsize\ttfamily}}{}

\newcommand{\RR}{\mathbb{R}}

\renewcommand{\d}[1]{\sfrac{d\!}{d #1}\ }
\newcommand{\dx}{\d{x}}

\newcommand{\du}{\d{u}}
\newcommand{\dv}{\d{v}}
\newcommand{\dd}[2]{\sfrac{d #1\!}{d #2}\ }

\newcommand{\norm}[1]{\left\Vert #1 \right\Vert}

\newcommand{\silent}[1]{}

\lstdefinelanguage{DOT}%
{morekeywords={val,new},%
  sensitive,%
  morecomment=[l]//,%
  morecomment=[s]{/*}{*/},%
  morestring=[b]",%
  morestring=[b]',%
  showstringspaces=false%
}[keywords,comments,strings]%

\newlength{\trulemargin}
\newlength{\trulewidth}
\newlength{\srulewidth}
\newenvironment{trules}{$\vspace{0.5em}\ba{p{\trulemargin}@{~}p{\trulewidth}@{~}p{\trulemargin}}}{\ea$}
\newenvironment{srules}{$\vspace{0.5em}\ba{p{\trulemargin}@{~}p{\srulewidth}}}{\ea$}

\newcommand{\at}{@}

\newcommand{\gap}{\quad\quad}

\newcommand{\ba}{\begin{array}}
\newcommand{\ea}{\end{array}}

\newcommand{\ei}{\end{array}}
\newcommand{\bcases}{\left\{\begin{array}{ll}}
\newcommand{\ecases}{\end{array}\right.}

\newcommand{\bra}[1]{\llbracket {#1} \rrbracket}
\newcommand{\bro}[0]{\overrightarrow{\mathcal{D}}}
\newcommand{\orb}[0]{\overleftarrow{\mathcal{D}}}

\graphicspath{{figures/}}

\setcopyright{rightsretained}
\acmPrice{}
\acmDOI{10.1145/3341700}
\acmYear{2019}
\copyrightyear{2019}
\acmJournal{PACMPL}
\acmVolume{3}
\acmNumber{ICFP}
\acmArticle{96}
\acmMonth{8}

\begin{document}

\title[Demystifying Differentiable Programming: Shift/Reset the Penultimate Backpropagator]{Demystifying Differentiable Programming: \\Shift/Reset the Penultimate Backpropagator}



\author{Fei Wang}
\affiliation{
  \institution{Purdue University}            
  \country{USA}
}

\author{Daniel Zheng}
\affiliation{
  \institution{Purdue University}            
  \country{USA}
}

\author{James Decker}
\affiliation{
  \institution{Purdue University}            
  \country{USA}
}

\author{Xilun Wu}
\affiliation{
  \institution{Purdue University}            
  \country{USA}
}

\author{Gr\'egory M. Essertel}
\affiliation{
  \institution{Purdue University}            
  \country{USA}
}

\author{Tiark Rompf}
\affiliation{
  \institution{Purdue University}            
  \country{USA}
}

\lstMakeShortInline[keywordstyle=,%
              flexiblecolumns=false,%
              mathescape=false,%
              basicstyle=\tt]@

\newcommand{\Let}[2]{\texttt{let}\ #1\ \texttt{=}\ #2\ \texttt{in} \ }
\newcommand{\SLet}[2]{\overline{\texttt{let}}\ #1\ \texttt{=}\ #2\ \overline{\texttt{in}} \ }
\newcommand{\DLet}[2]{\underline{\texttt{let}}\ #1\ \texttt{=}\ #2\ \underline{\texttt{in}} \ }
\newcommand{\WLet}[2]{\uwave{\texttt{let}}\ #1\ \texttt{=}\ #2\ \uwave{\texttt{in}} \ }
\newcommand{\LetRec}[2]{\texttt{letrec}\ #1\ \texttt{=}\ #2\ \texttt{in} \ }
\newcommand{\DLetRec}[2]{\underline{\texttt{letrec}}\ #1\ \texttt{=}\ #2\ \underline{\texttt{in}} \ }
\newcommand{\Ref}[1]{\texttt{ref} \ #1}
\newcommand{\DRef}[1]{\underline{\texttt{ref}} \ #1}
\newcommand{\Fst}[1]{\texttt{fst} \ #1}
\newcommand{\DFst}[1]{\underline{\texttt{fst}} \ #1}
\newcommand{\Snd}[1]{\texttt{snd} \ #1}
\newcommand{\DSnd}[1]{\underline{\texttt{snd}} \ #1}

\newcommand{\Left}[1]{\texttt{inl} \ #1 }
\newcommand{\DLeft}[1]{\underline{\texttt{inl}} \ #1}
\newcommand{\Right}[1]{\texttt{inr} \ #1 }
\newcommand{\DRight}[1]{\underline{\texttt{inr}} \ #1 }

\newcommand{\Pattern}[3]{\texttt{case} \ #1 \ \texttt{of} \ x^\tau \Rightarrow #2 \ \texttt{or} \ x^\tau \Rightarrow #3}
\newcommand{\Pat}[5]{\texttt{case} \ #1 \ \texttt{of} \ #2 \Rightarrow #3 \ \texttt{or} \ #4 \Rightarrow #5}
\newcommand{\DPat}[5]{\underline{\texttt{case}} \ #1 \ \underline{\texttt{of}} \ #2 \Rightarrow #3 \ \underline{\texttt{or}} \ #4 \Rightarrow #5}
\newcommand{\DPatL}[5]{\underline{\texttt{case}} \ #1 \\
                      &&\underline{\texttt{of}} \ #2 \Rightarrow #3 \\
                      &&\underline{\texttt{or}} \ #4 \Rightarrow #5}
\newcommand{\PatL}[5]{\texttt{case} \ #1 \\
                      &&\texttt{of} \ #2 \Rightarrow #3 \\
                      &&\texttt{or} \ #4 \Rightarrow #5}
\newcommand{\REF}[1]{\texttt{Ref} \ #1}
\newcommand{\DREF}[1]{\underline{\texttt{Ref}} \ #1}
\newcommand{\Dat}[0]{\underline{\at} \ }
\newcommand{\Sat}[0]{\overline{\at} \ }
\newcommand{\Wat}[0]{\uwave{\At} \ }
\newcommand{\At}[0]{\at \ }
\newcommand{\WLam}[0]{\uwave{\lambda\text{ }}}
\newcommand{\SLam}[0]{\overline{\lambda\text{ }}}
\newcommand{\DLam}[0]{\underline{\lambda\text{ }}}

\newcommand{\DAdd}[2]{\ #1 \ \underline{+} \ #2 \ }
\newcommand{\DMul}[2]{\ #1 \ \underline{*} \ #2 \ }
\newcommand{\DUpdate}[2]{#1 \ \underline{\mathrel{+}=} \ #2}
\newcommand{\DDref}[1]{\ \underline{!} \ #1}

\newcommand{\Shift}[2]{\texttt{shift} \ #1 \ \texttt{in} \ #2}
\newcommand{\SShift}[2]{\overline{\texttt{shift}} \ #1 \ \overline{\texttt{in}} \ #2}
\newcommand{\Res}[1]{\langle \ #1 \ \rangle}
\newcommand{\SRes}[1]{\overline \langle \ #1 \ \overline \rangle}
\newcommand{\DPair}[2]{\underline{(} #1 \ \underline{,} \ #2 \underline{)}}
\newcommand{\SPair}[2]{\overline{(} #1 \ \overline{,} \ #2 \overline{)}}

\newcommand{\UNIT}[0]{\texttt{Unit}}
\newcommand{\NUM}[0]{\RR \times (\REF \RR)}

\newcommand{\BOOL}[0]{\texttt{Boolean}}
\newcommand{\TR}[0]{\texttt{Tree}}
\newcommand{\TRR}[0]{\texttt{Tree'}}
\newcommand{\A}[0]{\texttt{A}}
\newcommand{\Leaf}[1]{\texttt{Leaf} \ #1}
\newcommand{\Branch}[2]{\texttt{Branch} \ #1 \ #2}

\newcommand{\Exp}[0]{\texttt{EXP}}
\newcommand{\Val}[0]{\texttt{VAL}}
\newcommand{\Id}[0]{\texttt{ID}}
\newcommand{\Env}[0]{\texttt{ENV}}

\newcommand{\Red}[1]{\textcolor[rgb]{0.00,0.00,1.00}{#1}} 
\newcommand{\Green}[1]{\textcolor[rgb]{0.00,1.00,0.00}{#1}}
\newcommand{\Blue}[1]{\textcolor[rgb]{0.00,0.00,1.00}{#1}}
\newcommand{\ifThenElse}[3]{\texttt{if} \ #1 \ \texttt{then} \ #2 \ \texttt{else} \ #3}

\begin{abstract}
Deep learning has seen tremendous success over the past decade in
computer vision, machine translation, and gameplay.
This success rests crucially on \emph{gradient-descent optimization}
and the ability to ``learn'' parameters of a neural network by
backpropagating observed errors. However, neural network architectures
are growing increasingly sophisticated and diverse, which motivates
an emerging quest for even more general forms of \emph{differentiable programming},
where arbitrary parameterized computations can be trained by gradient
descent.
In this paper, we take a fresh look at automatic differentiation (AD)
techniques, and especially aim to demystify the \emph{reverse-mode} form of AD
that generalizes backpropagation in neural networks.

We uncover a tight connection between reverse-mode AD and delimited continuations,
which permits implementing reverse-mode AD purely via operator overloading
and without managing any auxiliary data structures.
We further show how this formulation of AD can be fruitfully combined
with multi-stage programming (staging), leading to an efficient
implementation that combines the performance benefits of deep learning frameworks
based on explicit reified computation graphs (e.g., TensorFlow) with
the expressiveness of pure library approaches (e.g., PyTorch).

\end{abstract}



\maketitle

\section{Introduction}

Under the label \emph{deep learning}, artificial neural
networks have seen a remarkable renaissance over the last decade.
After a series of rapid advances, they now match or surpass
human performance in computer vision, machine translation, and gameplay.
Common to all these breakthroughs is the underlying dependency
on optimization by gradient descent: a neural network ``learns''
by adjusting its parameters in a direction that minimizes the
observed error on a task. Hence, a crucial ability is that
of backpropagating errors through the network to compute the
gradient of a loss function \cite{rumelhart1986learning}.
Beyond this commonality, however, deep learning architectures
vary widely.
In fact, many of the practical successes are fueled by
increasingly sophisticated and diverse network architectures
that in many cases depart from the traditional organization
into layers of artificial neurons. For this reason, prominent deep learning researchers
have called for a paradigm shift from deep learning towards
\emph{differentiable programming} \cite{blog:nn_fp, blog:differentiable_programming}
--- essentially, functional
programming with first-class gradients --- based on
the expectation that further advances in artificial intelligence
will be enabled by the ability to ``train'' arbitrary parameterized
computations by gradient descent.

Programming language designers and compiler writers, key players in this vision, are
faced with the challenge of adding efficient and
expressive program differentiation capabilities.
Forms of automatic gradient computation that generalize
the classic backpropagation algorithm are provided by all
contemporary deep learning frameworks, including TensorFlow
and PyTorch. These implementations, however, are ad-hoc,
and each framework comes with its own set of
trade-offs and restrictions.
In the academic world, automatic differentiation (AD)
\citep{DBLP:journals/cacm/Wengert64,speelpenning1980compiling}
is the subject of study of an entire community.
Unfortunately, results disseminate only slowly
between communities, and while the forward-mode flavor of AD
is easy to grasp, descriptions of the reverse-mode flavor that
generalizes backpropagation often appear
mysterious to PL researchers.
A notable exception is the seminal
work of \citet{DBLP:journals/toplas/PearlmutterS08}, which
cast AD in a functional programming framework and
laid the groundwork for first-class, unrestricted,
gradient operators in a functional language.
Recent work by \citet{DBLP:journals/pacmpl/Elliott18} presented
a unification of forward- and reverse-mode AD based on the
``compiling to categories'' approach \citep{DBLP:journals/pacmpl/Elliott17a},
translating Haskell code to parameterized cartesian closed categories.
However, the technique still needs primitive functor-level loop-style
operations such as @map@, @sum@, and @zip@, and currently lacks
support for general recursion or Turing-completeness.

The goal of the present work is to further demystify differentiable
programming and reverse-mode AD for a PL audience, and to reconstruct
the forward- and reverse-mode AD approaches
based on well-understood program transformation
techniques, without relying on category theory.
We describe forward-mode AD as the symbolic differentiation
of ANF-transformed programs, and reverse-mode AD as
a specific form of symbolic differentiation of
CPS-transformed programs.
In doing so, we uncover a deep connection between
reverse-mode AD and delimited continuations.

In contrast to previous descriptions,
this formulation suggests a novel view of reverse-mode AD
as a purely local program transformation which can be
realized entirely using operator overloading in a language
that supports @shift@/@reset@ \cite{DBLP:conf/lfp/DanvyF90}
or equivalent delimited control operators\footnote{Our description reinforces
the functional ``Lambda, the ultimate backpropagator'' view of
\citet{DBLP:journals/toplas/PearlmutterS08}
with an alternative encoding based on delimited continuations,
where control operators like \texttt{shift/reset} act as a powerful
front-end over $\lambda$-terms in CPS ---
hence, as the ``penultimate backpropagator''.}.
By contrast, previous descriptions require non-local
program transformations to carefully manage
auxiliary data structures
(often called a \emph{tape}, \emph{trace}, or \emph{Wengert-list}~\cite{DBLP:journals/cacm/Wengert64}),
either represented explicitly, or in a refunctionalized form
as in \citet{DBLP:journals/toplas/PearlmutterS08}.

Delimited control operators lead to an expressive implementation in the
(define-by-run) style of PyTorch. We further show how to combine this approach
with \emph{multi-stage programming} to derive a framework in the
(define-then-run) style of TensorFlow.
The result is a highly-efficient and expressive DSL,
dubbed Lantern\footnote{\url{https://github.com/feiwang3311/Lantern}\label{github:lantern}}, that reifies computation
graphs at runtime in the style of TensorFlow~\cite{DBLP:journals/corr/AbadiABBCCCDDDG16},
but also supports unrestricted control flow in the style of
PyTorch~\cite{paszke2017pytorch}.
Thus, our approach combines the strengths of these
systems without their respective weaknesses, and
explains the essence of deep learning frameworks
as the combination of two well-understood and
orthogonal ideas: staging and delimited continuations.

We first presented the idea of reverse-mode AD via delimited continuations and staging as a 
poster and accompanying abstract in the workshop track at ICLR \citep{wang2018a},
followed by a detailed tech-report on arXiv \citep{DBLP:journals/corr/abs-1803-10228}.
We then presented this idea to the Machine Learning community at NeurIPS
\citep{DBLP:conf/nips/FeiWang18}, along with an evaluation of our prototypic
implementation (which only supported a CPU backend)
of the framework Lantern. 
The NeurIPS paper focused primarily on the intuitions and high-level ideas; no formal presentation was provided.
The current paper presents a unified view of automatic differentiation from a PL
perspective and extends earlier publications through the following
contributions:
\medskip
\begin{itemize}[leftmargin=4mm]

\item We first bridge the conceptual distinction between automatic differentiation and symbolic differentiation
by casting forward-mode AD
as the application of standard high-school symbolic differentiation rules on ANF-transformed terms, with
only constant expression size increase. Based on that insight, we define a formal
transformation that implements forward-mode AD directly (Section~\ref{sec:forward}).

\item We then analyze reverse-mode AD, and relate its ``there and back again'' computation flow pattern
to programs using nested continuations, as seen in CPS (continuation-passing style).
By presenting detailed formal transformations
(available as artifact online\footnote{\url{https://github.com/feiwang3311/demystifying-ad}\label{github:supplemental}})
for reverse-mode AD based on CPS, with or without
the use of control operators (shift/reset), in the target or meta-language, we
reveal the formal relationship between reverse-mode AD and CPS transformation (Section~\ref{sec:reverse}).

\item We demonstrate different ways to combine our forward- and reverse-mode AD for higher order gradients, and
present a concrete OO-style class hierarchy for higher-order AD
(Sections~\ref{sec:perturb} \ref{sec:highorderforward} \ref{sec:highorderreverse}, available as artifact online\textsuperscript{\ref{github:supplemental}}).
We also discuss the question of mutability and describe one way to make reverse-mode AD purely functional
via store-passing using an immutable map data-structure (Section~\ref{sec:purelyfunc}).

\item We illustrate the interplay between CPS transformation and staging, and relate the
implementation of control flow operations (@IF@, @WHILE@, and @TREE@ as a representative of recursive
machine learning models) to formal rules for reverse-mode AD transformation.
We also demonstrate examples showing intermediate code generation steps (Section~\ref{sec:lms}).

\item We demonstrate the performance of the complete
Lantern framework on realistic benchmark models (TreeLSTM, SqueezeNet, ResNet, and DeepSpeech2) on GPU
(Section~\ref{sec:eval}).

\end{itemize}

\medskip\noindent
Finally, Section~\ref{sec:related} discusses related work, and
Section~\ref{sec:conclusions} offers concluding thoughts.
\section{Differentiable Programming Basics}\label{sec:forward}

Broadly speaking, a neural network is a specific kind of
parameterized function approximator $\hat{f}_{{w}}$.
The training process optimizes the parameters ${w}$ to improve the
approximation of an unknown \emph{ground truth} function $f$ based
on training data.
$$f: A \to B \gap \hat{f}_{{w}}: A \to B \gap w \in P$$

For training, we take input/output samples $(a, f(a)) \in A \times B$
and update ${w}$ according to a \emph{learning rule}.
In typical cases where the functions $f$ and $\hat{f}_w$ are maps $\mathbb{R}^n \rightarrow \mathbb{R}^m$ and $w$ is of the form $\mathbb{R}^k$,
we want to find
the weights ${w}$ that achieve the smallest error or loss
$L({w}) = \norm{f(a) - \hat{f}_{{w}}(a)}$ on a given training set,
in the hope that the training set is representative enough that
the quality of the approximation of $\hat{f}_w$ will
generalize to other inputs of $f$.

While there exists a myriad of ways to update ${w}$, the most popular
method is gradient descent. This is largely due to the fact
that gradients can be computed efficiently even for extremely large
numbers of parameters. We briefly describe gradient descent, as follows:

Given a training sample $(a, f(a)) \in A \times B$ and some initialization of ${w}$ at ${w^i}$,
both the loss $L({w}^i)$ and the gradient\footnote{The gradient $\nabla f$ of a function $f: \RR^n \to \RR$ is defined as the vector of partial derivatives of $f$ with respect to each of its parameters:
$\nabla f(u) = (\frac{\partial f(u)}{\partial u_1}, \frac{\partial f(u)}{\partial u_2}, ... \ , \frac{\partial f(u)}{\partial u_n})$}
$\nabla L({w^i})$ can be computed.
The gradient marks the direction which increases the loss $L({w}^i)$ the most
rapidly, and the gradient descent algorithm dictates that ${w}$ should
be updated in the direction of the negative gradient by a small step
proportional to the \emph{learning rate} $r$.
$$ {w}^{i+1} = {w}^i - r * \nabla L({w^i}) $$
This update step is performed many times.
In practice, however, gradient descent is almost never used in this
pure form. Most commonly used are \emph{stochastic gradient descent} (SGD) flavors
that operate on batches of training samples at a time. Popular variants are
SGD with momentum~\cite{DBLP:journals/nn/Qian99}, Adagrad~\cite{DBLP:journals/jmlr/DuchiHS11}, and Adam \cite{DBLP:journals/corr/KingmaB14}.

An important property of gradient computation is that differentiability is compositional.
Traditional neural networks (i.e., those organized into
layers) are simple function compositions
$\hat{f}_{{w}} = \hat{f}_{n,w_n} \circ \ldots \circ \hat{f}_{1,w_1}$
where each $\hat{f}_{i,w_i}$ represents a layer.
Other architectures compose in a similar way and enable end-to-end
training. A popular example is image captioning, which composes
convolutional neural networks (CNN) \cite{lecun1990handwritten} and
recurrent neural networks (RNN) \cite{elman1990finding}.

Imagine, however, that $\hat{f}_{{w}}$ and by extension $L(w)$ is not just a
simple sequence of function compositions, but is instead defined by
a \emph{program}, e.g., a $\lambda$-term with complex control flow.
How, then, should $\nabla L(w)$ be computed?

\subsection{From Symbolic Differentiation to Forward-Mode AD}\label{sec:symbo}

\begin{figure}
\begin{minipage}[t]{0.3\textwidth}
Syntax:
$$\small
\ba{rcl}
e &::= &c \\
&|& x \ \   \\
&|& e + e \ \\
&|& e * e  \   \\[1ex]
&|& \text{let} \ x = e \ \text{in} \ e
\ea
$$
\end{minipage}\hspace{1ex}%
\begin{minipage}[t]{0.6\textwidth}
Symbolic differentiation rules:
$${\small
\ba{rcl}
\dx \bra{c} &=& 0  \\
\dx \bra{x} &=& 1 \ \\
\dx \bra{e_1 + e_2} &=& \dx \bra{e_1} + \dx \bra{e_2} \ \\
\dx \bra{e_1 * e_2} &=& \dx \bra{e_1} * e_2 + e_1 * \dx \bra{e_2} \\[1ex]
\dx \bra{\text{let} \ y = e_1 \ \text{in} \ e_2} &=&
\text{let}\ y\ = e_1\ \text{in} \\
&&\text{let}\ y' = \dx \bra{e_1}\ \text{in} \\
&&\dx \bra{e_2}\\
\dx \bra{y} &=& y' \quad (y \ne x) \\
\ea}
$$
\end{minipage}
\caption{\label{fig:grammar1} Symbolic differentiation for a simple
expression language, extended with \texttt{let} expressions.}
\vspace{-3ex}
\end{figure}

Symbolic differentiation techniques to obtain the derivative of an expression
are taught in high schools around the world. Some of the most well-known rules
are shown in Figure~\ref{fig:grammar1} (the rule involving @let@ expressions is
explained shortly). As such, symbolic differentiation is the first candidate to
compute derivatives of program expressions.
However, some differentiation rules may cause code explosion; not
only in size, but also in terms of computation cost.
Consider the following example: \\
\begin{center}
\vspace{-3ex}
$\small
\ba{rll}
\dx \bra{e_1*e_2* ... *e_n} = &\dx \bra{e_1}*e_2* ... *e_n & + \\
&e_1*\dx \bra{e_2}* ... *e_n & + \\
&... & + \\
&e_1 * e_2 * ... * \dx \bra{e_n} \\
\ea$
\end{center}

The size-$n$ term on the left-hand side is transformed into $n$ size-$n$ terms,
which is a quadratic increase. Worse, each $e_i$ is now evaluated $n$ times.

This problem is well recognized in the AD community
and often cited as a major motivation for more efficient approaches.
In fact, many AD papers go to great lengths to explain that
``AD is not symbolic differentiation'' \citep{DBLP:journals/toplas/PearlmutterS08,DBLP:journals/corr/BaydinPR18}.
However, let us consider what happens if we convert the program to
administrative normal form (ANF) \cite{DBLP:conf/pldi/FlanaganSDF93} first,
binding each intermediate result in a @let@ expression: \\[-2ex]
\begin{minipage}[t]{0.3\textwidth}
$$\small
\ba{rll}
\dx \llbracket
&\text{let} \ y_1 = e_1\ \text{in}\\
&... \\
&\text{let} \ y_n = e_n\ \text{in}\\
&\text{let} \ z_1 = y_1 * y_2\ \text{in} \\
&\text{let} \ z_2 = z_1 * y_3\ \text{in} \\
&... \\
&\text{let} \ z_{n-1} = z_{n-2} * y_n\ \text{in} \\
&z_{n-1} \ \rrbracket \\
\ea$$
\end{minipage}
\begin{minipage}[t]{0.5\textwidth}
$$\small
\ba{rll} =
&\text{let} \ y_1 = e_1\ &\text{in}\ \text{let} \ y_1' = \dx \llbracket e_1 \rrbracket \ \text{in}\\
&... \\
&\text{let} \ y_n = e_n\ &\text{in}\ \text{let} \ y_n' = \dx \llbracket e_n \rrbracket\ \text{in}\\
&\text{let} \ z_1 = y_1 * y_2\ &\text{in}\ \text{let}\ z_1' = y_1' * y_2 + y_1 * y_2' \ \text{in} \\
&\text{let} \ z_2 = z_1 * y_3\ &\text{in}\ \text{let}\ z_2' = z_1' * y_3 + z_1 * y_3' \ \text{in} \\
&... \\
&\text{let} \ z_{n-1} = z_{n-2} * y_n\ &\text{in}\ \text{let}\ z_{n-1}' = z_{n-2}' * y_n + z_{n-2} * y_n' \\
&\text{in} \ z_{n-1}'
\ea$$
\end{minipage}\\[2ex]

After ANF-conversion, the expression size increases only by a constant factor.
The program structure remains intact, and just acquires an additional @let@
binding for each existing binding. No expression is evaluated more often
than in the original computation.

This example uses the standard symbolic differentiation rules for addition and multiplication,
but also makes key use of the @let@ rule in Figure~\ref{fig:grammar1},
which splits a binding @let@~$y = ...$ into @let@~$y = ...$ and @let@~$y' = ...$.
Using terminology from the AD community, we call $y$ the \emph{primal} and
$y'$ the \emph{tangent}. The rules in Figure~\ref{fig:grammar1} work with respect to a fixed $x$,
which we assume by convention does not occur bound in any $\text{let} \ x = ...$
expression. All expressions are of type $\RR$, so a derivative can be computed for
any expression. We write $\dx \bra{e}$ using bracket syntax to emphasize that
symbolic differentiation is a syntactic transformation.

Symbolic differentiation of ANF-transformed terms maintains the asymptotic runtime-complexity.
Let us consider a concrete example: $y = 2 * x + x * x * x$.
We start from its ANF-transformed form.\\
\begin{minipage}{0.26\textwidth}
\vspace{-1ex}
$$\small
\ba{rll}
\dx\llbracket
&\Let{y_1}{2 * x}\\
&\Let{y_2}{x * x}\\
&\Let{y_3}{y_2 * x}\\
&\Let{y}{y_1 + y_3}\\
&y \ \rrbracket \\
\ea$$
\end{minipage}
\begin{minipage}{0.68\textwidth}
$$\small
\ba{rll}
=
&\Let{y_1}{2 * x} \ &\Let{y_1'}{2 * x'} \\
&\Let{y_2}{x * x} \ &\Let{y_2'}{x' * x + x * x'} \\
&\Let{y_3}{y_2 * x} \ &\Let{y_3'}{y_2' * x + y_2 * x'} \\
&\Let{y}{y_1 + y_3} \ &\Let{y'}{y_1' + y_3'}\\
&y' \\
\ea$$
\end{minipage}

Note that we differentiate with respect to $x$, and $x' = 1$.
The computation of derivatives follows the rules in Figure~\ref{fig:grammar1}.
The final two @let@ bindings compute the primal value $y$ and the tangent $y'$ of
$y$ with respect to $x$.
The tangent is returned as the result of the program after transformation.
We can confirm the correctness of the calculation easily, which reduces to $2 + 3 * x * x$.

\begin{figure}
\vspace{-3ex}
\begin{minipage}{0.65\textwidth}
\centering
$$\small
\ba{l}
\text{We denote } p_{tk} \text{ as the } k \text{th parameter of } t \text{th computation,}\\
\text{where } p_{tk} \in \{c\} \cup \{x\} \cup \{y_j | j < t\}\\
\\
\Let{y_1}{p_{11} \oplus p_{12}} \\
\Let{y_1'}{\d{p_{11}}{\bra{p_{11} \oplus p_{12}}}\ *\ p_{12}' + \d{p_{12}}{\bra{p_{11} \oplus p_{12}}}\ *\ p_{11}'}     \\
\Let{y_2}{p_{21} \oplus p_{22}} \\
\Let{y_2'}{\d{p_{21}}{\bra{p_{21} \oplus p_{22}}}\ *\ p_{22}' + \d{p_{22}}{\bra{p_{21} \oplus p_{22}}}\ *\ p_{21}'}     \\
...                 \\
\Let{y_n}{p_{n1} \oplus p_{n2}} \\
\Let{y_n'}{\d{p_{n1}}{\bra{p_{n1} \oplus p_{n2}}}\ *\ p_{n2}' + \d{p_{n2}}{\bra{p_{n1} \oplus p_{n2}}}\ *\ p_{n1}'}     \\
y_n' \\
\ea
$$
\end{minipage}
\begin{minipage}{0.30\textwidth}
\centering
  \includegraphics[height=45mm]{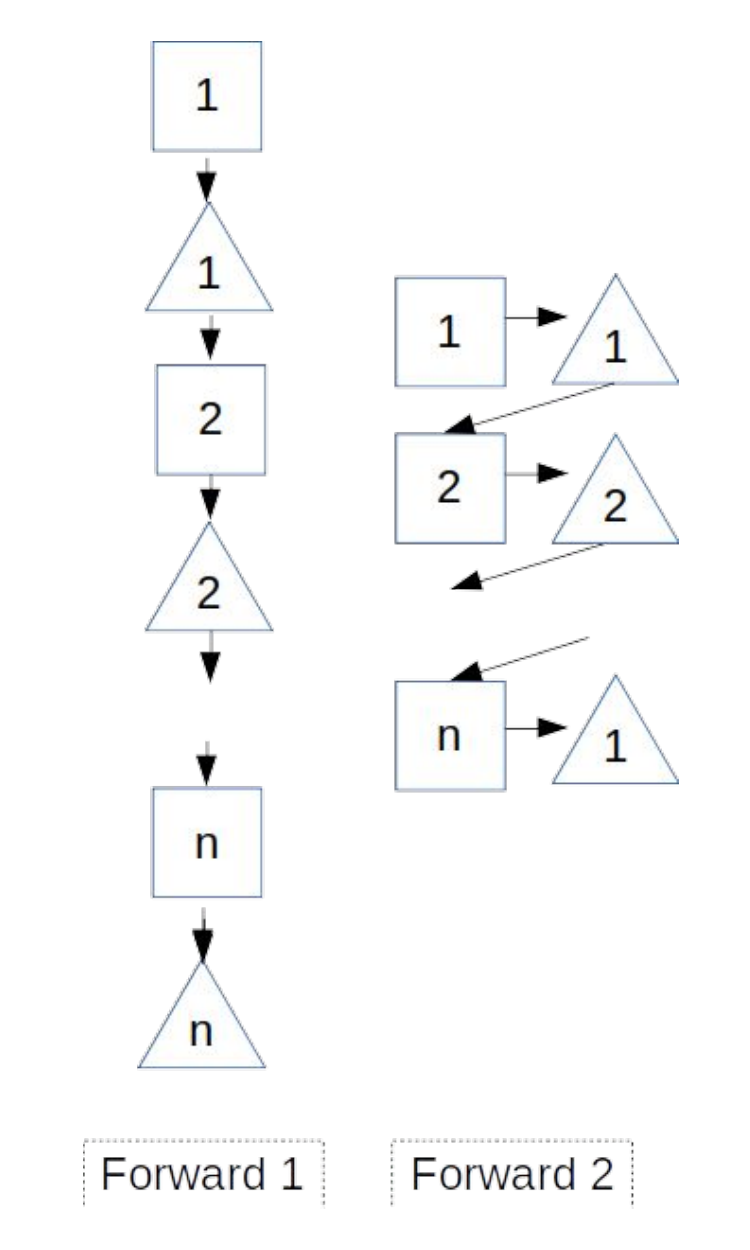}
\end{minipage}
\vspace{-3ex}
\caption{Pattern of computation of forward-mode AD for generic straight-line program}\label{fig:forward-generic}
\vspace{-3ex}
\end{figure}

For a generic straight-line program,
we can see this pattern of computation of forward-mode AD (Figure ~\ref{fig:forward-generic}).
The abstract flow of the forward-mode AD is depicted in Figure ~\ref{fig:forward-generic} on the right.
We use squares to denote value computations, and triangles to denote gradient computations.
The transformed program interleaves value computations with gradient computations (Forward 1).
We can further combine each value computation with its gradient computation (Forward 2),
so that the transformation can be realized via operator overloading.

For straight-line programs, applying ANF conversion followed by symbolic differentiation
achieves exactly the standard presentations of forward-mode AD.
Hence, it seems to us that the AD community has taken a too narrow view of
symbolic differentiation, excluding the possibility of @let@ bindings, and we
believe that repeating the mantra ``AD is not symbolic differentiation''
is ultimately harmful and contributes to the mystical appearance of the field.
We believe that understanding sophisticated AD algorithms
as \emph{specific forms} of symbolic differentiation will overall lead
to a better understanding of these techniques.

\vspace{-1ex}

\subsection{Forward-Mode AD for Lambda Calculus}

\begin{figure}[h]
\vspace{-1ex}
$$\footnotesize
\ba{rll}
\textbf{CORE LANGUAGE} \\
\textbf{Expressions:} && \Exp\\
e &::=& c \ | \  x \  | \ e \ + \ e  \ |\  e \ * \ e \ | \ \lambda x. \ e \ | \ \At e \ e \ | \ \texttt{let} \ x \ = \ e \ \texttt{in} \ e \\
    &|& \texttt{fst} \ e \ | \ \texttt{snd} \ e \ | \ ( e, \ e ) \ | \ \texttt{ref} \ e \ | \ ! \ e \ | \ e := e \\
    &|& \Left{e} \ | \ \Right{e} \ | \ \Pat{e}{x}{e}{x}{e} \\
    &|& \Shift{x}{e} \ | \ \Res{e} \\
\textbf{Values:} && \Val\\
                 && \text{structure left abstract}\\
\\
\textbf{DERIVED CONSTRUCTS} \\
\textbf{Booleans and conditionals:} \\
\text{Value: True}   &=& \Left{()}  \\
\text{Value: False}  &=& \Right{()} \\
\ifThenElse{b}{t}{e} &=& \Pat{b}{y}{t}{z}{e} \\
\textbf{Loops and recursion:} \\
\LetRec{f}{\lambda x. \ e_1} e_2 &=& \Let{f_0}{\lambda f_1. \lambda x. \ \Let{f}{\At f_1 \ f_1} e_1}\\
                                  && \Let{f}{\At f_0 \ f_0} e_2\\
                \text{Loops:} && \text{expressed as tail recursive functions}\\
\textbf{Tree data structures:} \\
\text{Example tree term: } t &=& \Right{(\Left{5}, \Left{6})}\\
\\
\textbf{Syntactic sugar:} \\
    y_1 \ \mathrel{+}= \ ! \ y_2 &=& y_1 \ := \ ! \ y_1 \ + \ ! \ y_2 \\
    \Let{(y, y')}{e_1}{e_2} &=& \Let{\tilde y}{e_1} \Let{y}{\Fst \tilde y} \Let{y'}{\Snd \tilde y} e_2 \\
    e_1 \ ; \ e_2 &=& \Let{\_}{e_1} e_2 \\
\\
\ea
$$\vspace{-5ex}
\caption{
  Formal definition of the language we consider. It serves as both object- and meta-language (for transformation).
  We show the syntax of the core languages (untyped, but types can be added),
  as well as derived constructs that express branches,
  loops, recursion, and recursive data structures in a standard way.
  Syntactic sugar used in our presentation is also listed here.\\[1ex]
  We assume Barendregt's variable convention throughout, such that all bound variables are pairwise different
  and different from the free variables.
  This allows several rules
  to be simplified compared to other formulations
  (no need for variable substitutions in transformations). \\[1ex]
  For transformation, we assume that the target language is the same as the object language unless noted otherwise.
}
\label{fig:formal}
\vspace{-3ex}
\end{figure}

\begin{footnotesize}
\begin{figure}[h]
\vspace{-2ex}
\flushleft{
For automatic differentiation (both forward-mode and reverse-mode in later sections),
we use the following variable sugaring $\hat .$ notation.
This variable sugaring is not strictly necessary but we find it convenient
for $+$ and $*$ rules. Also, note that this variable sugaring is always used
at positions where we know for sure that the
sugared variables bind with $\RR$ typed values, so that they must have gradients
(denoted via variables with $'$).
}
$$
\ba{rll}
\textbf{Variable Sugaring:} \quad \quad \hat y &=& (y, y') \\
\ea
$$

Also note that for AD (both forward-mode and reverse-mode), we drop @shift@/@reset@ terms
from the source language, since the focus is to provide a semantics for AD in a
standard language, and @shift@/@reset@ will play a crucial role for the semantics
of AD transformation in reverse mode. Our AD also supports mutable state in the source language.

$$
\ba{rll}
\texttt{Transform}(f) &=& \lambda x. \ \Let{\hat y}{\At \bro\bra{f} \ (x, 1)} \ y' \\
\texttt{where} \ \bro \bra{.}: && \Exp \rightarrow \Exp \ \ \ \text{is defined as below:} \\
\\
\bro \bra{c}           &=& c  \ \ \text{if} \ c \notin \RR\\
\bro \bra{c}           &=& \Red{(c, 0)} \ \ \text{if} \ c \in \RR\\
\bro \bra{y}            &=& y \\

\bro \bra{e_1 + e_2} &=&
        \Blue{\Let{\hat y_1}{\bro\bra{e_1}}}\\
\ \ \ &&\Blue{\Let{\hat y_2}{\bro\bra{e_2}}}\\
\ \ \ &&\Blue{(y_1 + y_2, y_1' + y_2')} \\
\bro \bra{e_1 * e_2} &=&
        \Blue{\Let{\hat y_1}{\bro\bra{e_1}}} \\
\ \ \ &&\Blue{\Let{\hat y_2}{\bro\bra{e_2}}} \\
\ \ \ &&\Blue{(y_1 * y_2, y_1 * y_2' + y_1' * y_2)} \\
\bro \bra{\lambda y. \ e} &=& \lambda y. \ \bro \bra{e} \\
\bro \bra{\At e_1 \ e_2} &=& \At \bro\bra{e_1} \ \bro\bra{e_2} \\
\bro \bra{\text{let} \ y = e_1 \ \text{in} \ e_2} &=&
 \Let{y}{\bro\bra{e_1}} \bro\bra{e_2} \\
\bro\bra{\Fst e} &=& \Fst{\bro\bra e} \\
\bro\bra{\Snd e} &=& \Snd{\bro\bra e} \\
\bro\bra{\Ref e} &=& \Ref{\bro\bra e} \\
\bro\bra{! \ e}  &=& \ ! \ \bro\bra{e} \\
\bro\bra{e_1 := e_2} &=& \bro\bra{e_1} := \bro\bra{e_2} \\
\bro\bra{(e_1, e_2)} &=& (\bro\bra{e_1}, \bro\bra{e_2}) \\
\bro\bra{\Left{e}} &=& \Left{\bro\bra{e}} \\
\bro\bra{\Right{e}}&=& \Right{\bro\bra{e}} \\
\bro\bra{\Pat{e}{y_1}{e_1}{y_2}{e_2}} &=&
    \Pat{\bro\bra{e}}{y_1}{\bro\bra{e_1}}{y_2}{\bro\bra{e_2}} \\
\ea
$$
\vspace{-3ex}
\caption{
Transformation rules for forward-mode AD.
Note that there is no metalanguage redex generated in the transformation,
so by default, all constructs on the right-hand-sides are dynamic/target language constructs.
Rules that are different from the standard are highlighted in blue.
}
\label{fig:formal-forward}
\vspace{-4ex}
\end{figure}
\end{footnotesize}

We now proceed beyond straight-line programs and formalize a variant of
$\lambda$-calculus with @let@ bindings, products, sum-type constructs
($\texttt{inl}$, $\texttt{inr}$, $\texttt{case}$), and mutable state (Figure~\ref{fig:formal}).
The language also contains delimited control operators shift and reset (denoted via $\Res{.}$),
which will be used in later sections.
Note that the language is untyped, though types can be added in a standard way.
Control operators (@shift@/@reset@) and mutable state are orthogonal features,
so their interaction does not pose any difficulties.

We define a new differentiation operator $\bro \bra{e}$, where the arrow indicates forward-mode,
and provide the forward-mode AD transformation rules in Figure~\ref{fig:formal-forward}.
Note that differentiation is still with respect to a fixed $x$.
However, we always transform abstractions
(for any non-abstraction term $e$, we add an $\eta$-redex, and perform $\At \bro \bra{\lambda x. e} \ (x, 1)$).
By Barendregt's variable convention, $\bro\bra{.}$
never applies to the special variable $x$, thus the $\bro\bra{x}$ rule is elided in the formal presentation.

Compared to Section~\ref{sec:symbo}, we no longer rely on an ANF-pre-transform pass.
Instead, the rules for addition and multiplication insert @let@ bindings
directly. It is important to note that the resulting program may not be in ANF
due to nested @let@ bindings, but code duplication is still eliminated due
to the strict pairing of primals and tangents.
Readers acquainted with forward-mode AD will note that this methodology is
standard \cite{DBLP:journals/corr/BaydinPR18}, though the presentation is not.

\vspace{-1ex}
\subsection{Implementation using Operator Overloading}

Pairing the primal and tangent values for numeric expressions handles
computations in different scopes easily because, in function applications, the
@let@ insertions require both the primal and tangent of the parameters to
perform the tangent computation.
Since the transformation is purely local, working with pairs of numeric
expressions makes it immediately clear that this strategy can be implemented
easily in standard programming languages by operator overloading. This is
standard practice, which we illustrate through our implementation in Scala
(Figure~\ref{fig:numf}).

\begin{figure}[h!]
\vspace{-5ex}
\begin{multicols}{2}
\begin{lstlisting}[language=Scala,basicstyle=\footnotesize\ttfamily]
// Differentiable number type.
class NumF(val x: Double, val d: Double) {
  def +(that: NumF) =
    new NumF(this.x + that.x, this.d + that.d)
  def *(that: NumF) =
    new NumF(this.x * that.x,
             this.d * that.x + that.d * this.x)
  ...
}
\end{lstlisting}
\columnbreak
\begin{lstlisting}[language=Scala,basicstyle=\footnotesize\ttfamily]
// Differentiation operator.
def grad(f: NumF => NumF)(x: Double) = {
  val y = f(new NumF(x, 1.0))
  y.d
}
// Example and test.
val df = grad(x => 2*x + x*x*x)
forAll { x => df(x) == 2 + 3*x*x }
\end{lstlisting}
\end{multicols}
\vspace{-4ex}
\caption{\label{fig:numf}Forward-mode AD in Scala (operator overloading)}
\vspace{-3ex}
\end{figure}

The @NumF@ class encapsulates the primal as @x@ and the tangent as @d@, with
arithmetic operators overloaded to compute primal and tangent values at the same time.
To use the forward-mode AD implementation, we still need to define an operator @grad@
to compute the derivative of any function @NumF => NumF@ (Figure~\ref{fig:numf} upper right).
Internally, @grad@ invokes its argument function with a tangent value of 1
and returns the tangent field of the function result.
In line with the previous sections, we only handle scalar functions, but the
approach generalizes to multidimensional functions as well.
An example using the @grad@ operator is shown in Figure~\ref{fig:numf} lower right.
Note that the constant @2@ is implicitly converted to @new NumF(2.0, 0.0)@
(tangents of constants are 0.0 because constants do not change).
The use of @Double@ instead of a generic number type is simply for clarity of presentation.
Note how the implementations in Figure \ref{fig:numf}
correspond directly to the formal rules in Figure \ref{fig:formal-forward}.

\subsection{Nested Gradient Invocation and Perturbation Confusion}\label{sec:perturb}

In the current implementation, we can compute the gradient of any function of type @NumF => NumF@
with respect to any given value using forward-mode AD. However, our @grad@ function is not truly
first-class, since we cannot apply it in a nested fashion, as in @grad(grad(f))@.
This prevents us from computing higher order derivatives, and from solving nested min/max problems
in the form of:
\vspace{-1ex}
$$\text{min}_x \text{max}_y f(x, y)$$

Yet, even this somewhat restricted operator has a few
subtleties.
There is a common issue with functional implementations of AD that,
like ours, expose a gradient operator within the language.
In the simple example shown below, the inner call to @grad@ should return 1,
meaning that the outer @grad@ should also return 1.

\begin{center}
\begin{lstlisting}[language=Scala,basicstyle=\footnotesize\ttfamily]
grad { x: NumF =>
  val shouldBeOne = grad(y => x + y)(1) // Evaluates to 2 instead of 1! Unexpected.
  val z = NumF(shouldBeOne, 0)
  x * z
}(1)
\end{lstlisting}
\end{center}

However, this is not what happens. The inner @grad@
function will also collect the tangent from @x@,
thus returning 2 as the gradient of @y@. The outer @grad@ will then give a
result of 2 as the gradient of @x@.
This issue is called \emph{perturbation confusion} because
the @grad@ function is confusing the perturbation (i.e. derivative)
of a free variable used within the closure with the perturbation
of its own parameter.

The root of this problem is that the two @grad@ invocations differentiate
with respect to different variables (outer @grad@ wrt. @x@, inner @grad@ wrt. @y@), and that
their gradient updates should not be mixed.
We do not provide any new solutions for perturbation confusion,
but our implementation can be easily extended to support known solutions,
either based on \emph{dynamic tagging} or based on types as realized in
Haskell\footnote{\url{http://conway.rutgers.edu/~ccshan/wiki/blog/posts/Differentiation}},
which lifts tags into the type system using rank-2 polymorphism,
just like the @ST@ monad \cite{DBLP:conf/pldi/LaunchburyJ94}.

\vspace{-1ex}
\subsection{First-Class Gradient Operator}\label{sec:highorderforward}

While not the main focus of our work, we outline one way in which our
@NumF@ definition can be changed to support first-class gradient
computation, while preventing perturbation confusion.
Inspired by DiffSharp \cite{DBLP:journals/corr/BaydinPS16}, we
change the class signatures as shown below.
We unify @NumF@ and @Double@ in the same abstract class @Num@,
and add a dynamic tag value @tag@.
The @grad@ operator needs to assign a new tag for each invocation,
and overloaded operators need to take tags into account
to avoid confusing different ongoing invocations of @grad@.

\begin{center}
\begin{lstlisting}[language=Scala,basicstyle=\footnotesize\ttfamily]
abstract class Num
class NumV(val x: Double) extends Num
class NumF(val x: Num, val d: Num, val tag: Int) extends Num {...}
def grad(f: Num => Num)(x: Num): Num = {...}
\end{lstlisting}
\end{center}

This class hierarchy provides a flexible way to compose higher-order gradient computation
(implementation available online\textsuperscript{\ref{github:supplemental}}).
Alternative implementations that use parametric types and type
classes instead of OO-style inheritance are also possible.

This concludes the core ideas of forward-mode AD.
Implementations based on operator overloading are simple and direct,
and exist in many languages.
As noted earlier, we propose that forward-mode AD be viewed as a
specific kind of symbolic differentiation, either using standard differentiation
rules after ANF-conversion, or using transformation rules that insert
@let@ bindings on the fly, operating on value-derivative pairs
(i.e. primals and tangents).

\section{Differentiable Programming with Reverse-Mode AD}\label{sec:reverse}

Forward-mode AD is straightforward to implement and generalizes to functions with
multiple inputs and outputs. However, it is inefficient for functions with many
inputs, and neural networks generally have many inputs and few outputs.
To compute the gradient of a function $f\!:\!\RR^n \to \RR$, we have to compute
$n$ forward derivatives either sequentially or simultaneously, but this leads to
$O(n)$ more operations than the original function. Is there a better approach?

We consider again $f\!:\RR^n \to \RR$ represented as a straight-line program in ANF, i.e., as a
sequence of $\texttt{let}\ y_j = e_j$ expressions, with inputs $x_i$ and output $y_m$.
The basic intuition is: instead of computing all $n * m$ internal derivatives
$\d{x_i} y_j$ as in forward-mode, we would rather only compute the $m+n$ derivatives
$\d{y_j} y_m$ and $\d{x_i} y_m$. For this, we need a way to compute
derivatives starting with $\d{y_m} y_m = 1$, and accumulate derivatives backwards through
the program until we reach the inputs $x_i$. This form of AD is called reverse-mode AD,
and is the basis for backpropagation for neural networks.
The approach generalizes to functions $\RR^n \to \RR^m$ with multiple outputs,
and is generally more efficient than forward-mode AD when $n >> m$.

But how do the gradients propagate backward? The basic idea is rooted in the
\emph{chain rule of differentiation}, which states that:
$$\small \du f(g(u)) = \dv f(v) * \du g(u) \ \text{where } v = g(u)$$
To interpret the chain rule in English, it says that the ``sensitivity''
of $f(g(u))$ to changes in $u$ is the ``sensitivity'' of $f(v)$ to changes in $v$, where $v = g(u)$,
amplified by the ``sensitivity'' of $g(u)$ to changes in $u$.

For the $e_1 * e_2$ term in the grammars in Figure~\ref{fig:grammar1}, we may be tempted
to write:
$$
\ba{ll}
\dd{y}{\bra{e_1}} &= \dd{y}{\bra{e_1 * e_2}} * \dd{\bra{u * e_2}}{u}\\
\dd{y}{\bra{e_2}} &= \dd{y}{\bra{e_1 * e_2}} * \dd{\bra{e_1 * u}}{u}, \ \text{where } u \ \text{is fresh variable} \\
\ea
$$

The rules can be read as: the ``sensitivity'' of $y$ to $\bra{e_1}$ is the ``sensitivity'' of $y$
to $\bra{e_1 * e_2}$ amplified by the
``sensitivity'' of $\bra{e_1}$'s \emph{context} to $\bra{e_1}$,
and the ``sensitivity'' of $y$ to $\bra{e_2}$ is the ``sensitivity'' of $y$ to $\bra{e_1 * e_2}$
amplified by the
``sensitivity'' of $\bra{e_2}$'s \emph{context} to $\bra{e_2}$,
For direct correlation between
the above grammars and the chain rule, just do the following substitutions in the first transformation rule:
the first $y$ to $f(g(u))$, $\bra{e_1}$ to $u$,
the second $y$ to $f(v)$,   $\bra{e_1 * e_2}$ to $v$, and $\bra{u * e_2}$ to $g(u)$.

\begin{figure}[h]
\vspace{-2ex}
\begin{footnotesize}
\begin{minipage}{0.65\textwidth}
$$
\ba{rll}
\text{Simple running example:} \ x \text{ is input} \\
y = 2 * x + x * x * x \ \text{where}\ x' = \Ref 0 \\
\textbf{Forward pass:} \quad \quad \quad             &&  \quad \quad \quad \textbf{Backward pass:} \\
\Let{(y_1, y_1')}{(2 * x, \ \Ref 0)}     \ \Downarrow  && \Uparrow \  x' \mathrel{+}= \ ! y_1' * 2;  \\
\Let{(y_2, y_2')}{(x * x, \ \Ref 0)}     \ \Downarrow  && \Uparrow \  x' \mathrel{+}= \ ! y_2' * x; \ x' \mathrel{+}= \ ! y_2' * x; \\
\Let{(y_3, y_3')}{(y_2 * x,\  \Ref 0)}   \ \Downarrow  && \Uparrow \  y_2' \mathrel{+}=\ ! y_3' * x;\  x' \mathrel{+}= \ ! y_3' * y_2; \\
\Let{(y_4, y_4')}{(y_1 + y_3,\  \Ref 0)} \ \Downarrow  && \Uparrow \ y_1' \mathrel{+}=\ ! y_4';\  y_3' \mathrel{+}=\ ! y_4'; \\
                                y_4'   &:=& 1.0; \\
\ea
$$
\end{minipage}
\begin{minipage}{0.30\textwidth}
\centering
  \includegraphics[height=35mm]{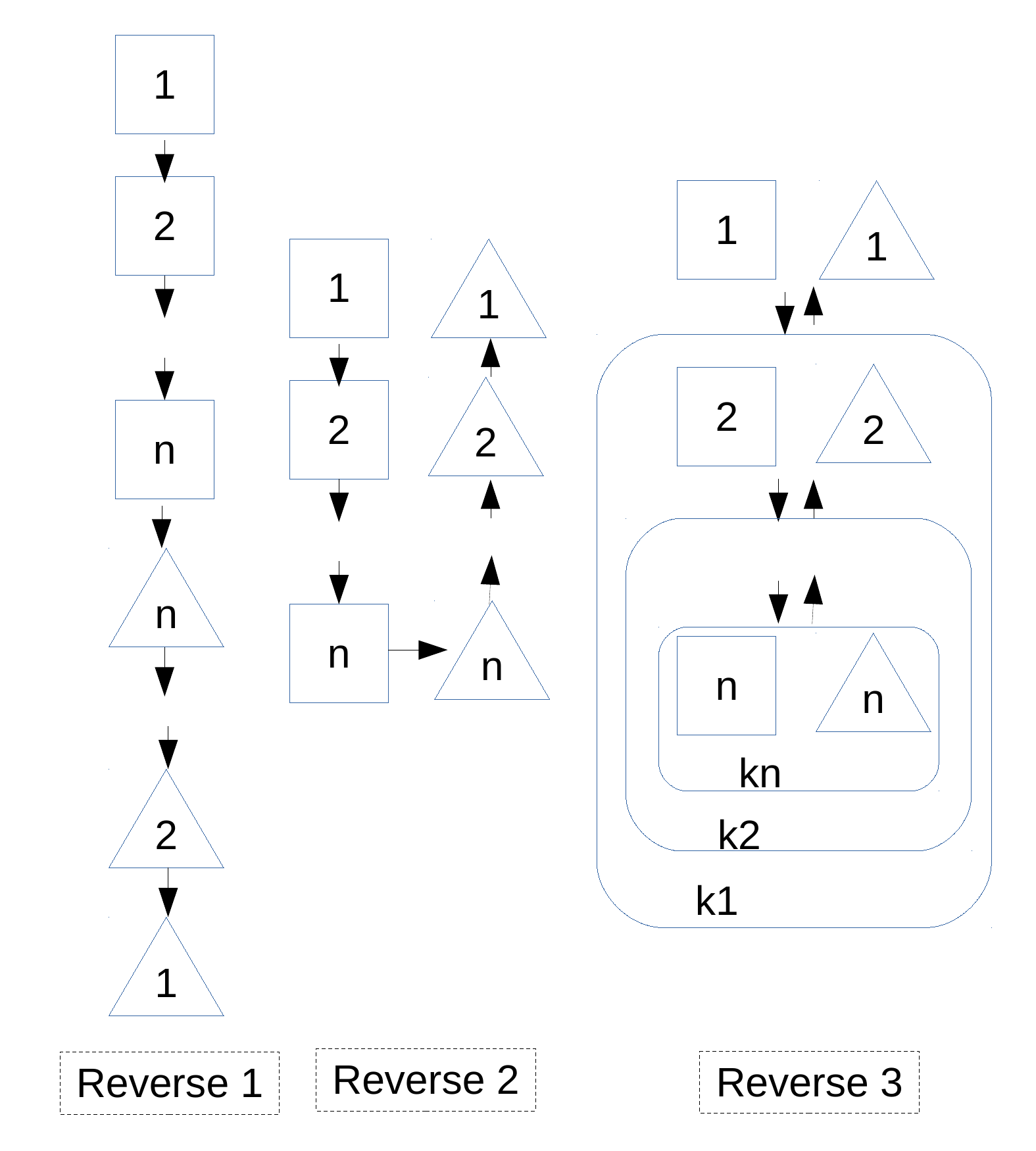}
\end{minipage}
\begin{minipage}{0.99\textwidth}
$$
\ba{ll}
&\text{General straight-line example:} \ x \text{ is input and } x' = \Ref 0 \\
&\text{We denote } p_{tk} \text{ as the } k \text{th parameter of } t \text{th computation, where } p_{tk} \in \{c\} \cup \{x\} \cup \{y_j | j < t\} \\
\ea
$$
$$
\ba{rll}
\textbf{Forward pass:}  \quad \quad \quad \quad                && \quad \quad \quad \quad \textbf{Backward pass:} \\
\Let{(y_1, y_1')}{(p_{11} \oplus p_{12}, \Ref 0)} \ \Downarrow && \Uparrow p_{11}' \mathrel{+}= \d{p_{11}}{\bra{p_{11} \oplus p_{12}}}\ *\ !y_1'; \
                                                                           p_{12}' \mathrel{+}= \d{p_{12}}{\bra{p_{11} \oplus p_{12}}}\ *\ !y_1';\\
\Let{(y_2, y_2')}{(p_{21} \oplus p_{22}, \Ref 0)} \ \Downarrow && \Uparrow p_{21}' \mathrel{+}= \d{p_{21}}{\bra{p_{21} \oplus p_{22}}}\ *\ !y_2'; \
                                                                           p_{22}' \mathrel{+}= \d{p_{22}}{\bra{p_{21} \oplus p_{22}}}\ *\ !y_2';\\
...                                               \ \Downarrow && \Uparrow ...                            \\
\Let{(y_n, y_n')}{(p_{n1} \oplus p_{n2}, \Ref 0)} \ \Downarrow && \Uparrow p_{n1}' \mathrel{+}= \d{p_{n1}}{\bra{p_{n1} \oplus p_{n2}}}\ *\ !y_n'; \
                                                                           p_{n2}' \mathrel{+}= \d{p_{n2}}{\bra{p_{n1} \oplus p_{n2}}}\ *\ !y_n';\\
                                                      y_n'    &:=&     1.0;   \\
\ea
$$
\end{minipage}
\end{footnotesize}
\vspace{-2ex}
\caption{Running examples of reverse-mode AD, the transformation of general straight-line programs,
and abstract computation flow that motivated continuation-passing style.}
\label{fig:reverse1}
\vspace{-1ex}
\end{figure}

However, the above transformation rules are not exactly correct. If both $e_1$ and $e_2$
contains $x$, then the sensitivity of $y$ to $x$ should be the sum of $\dd{y}{\bra{x}}$s that occurred
in multiple places (in both $\dd{y}{\bra{e_1}}$ and $\dd{y}{\bra{e_2}}$). This accumulation of gradients
is often modeled by mutable references and $\mathrel{+}=$ operations on the mutable references.
We call this \emph{destination-passing style},
where the reference cells accumulating the gradients are passed to the operations in the backward pass.
(An alternative pure functional implementation is discussed in Section~\ref{sec:purelyfunc}.)
Let us again try our running example $y = 2 * x + x * x * x$ (after ANF-transformation),
and map out the procedures for reverse-mode AD (Figure~\ref{fig:reverse1} top left).
Note that we have to run a \emph{forward-pass} first to compute and remember intermediate values, and then
a \emph{backward-pass} to accumulate the gradients. This is simply due to the fact that the derivative of the ``contexts''
may depend on the intermediate values computed in the forward pass.

For the running example, we deliberately reverse the statements in the backward-pass
(computation flow should follow the double arrows in the figure),
so that the forward-pass and the backward-pass of the same computation are on the same row.
A more general presentation of reverse-mode AD for straight-line programs is given in Figure~\ref{fig:reverse1} bottom.

Now if we look at the abstract computation flow shown in Figure~\ref{fig:reverse1} top right,
in comparison with forward-mode AD, the computation flow of reverse-mode AD processes all value computations
in the forward order, then processes all gradient computations in the reverse order (Reverse 1).
We can ``fold'' the gradient calculations up in parallel with the value calculations (Reverse 2),
like in our examples.
We can further nest the computations into continuations ($k_1$, $k_2$, ... , $k_n$ in Reverse 3),
following the inspiration from
``There and Back again'' \cite{danvy2005there}, and look for ways to model the
computation as a sequence of function calls, where the call path implements the forward
pass and the return path implements the backward pass.

With this intuition, it is not hard to see that a transformation to continuation-passing
style (CPS) provides exactly the right structure, i.e., for each computation step,
the subsequent forward-backward combinations
are contained in a set of nested continuations
(as $k_1, k_2, ... , k_n$ in Reverse 3, Figure~\ref{fig:reverse1} upper-right).
In contrast to regular CPS, our continuations do return
and can be followed by other computation/program statements. This kind of continuations,
which behaves more like regular callback functions, is called \emph{delimited continuations} \cite{DBLP:conf/popl/Felleisen88}.

\subsection{Implementation Using Operator Overloading}
We first express the idea via an implementation in Scala that
directly follows the intuitions in Figure~\ref{fig:reverse1}, where each overloaded operator
is provided with a (delimited) continuation $k$.
The code is shown in Figure~\ref{fig:CPS1}.
Just like in forward-mode AD, we associate values and their gradients as two fields of a
class, here @NumR0@.
Every operator takes a delimited continuation @k@, which is expected to take the intermediate variable @y@,
handle the rest of the forward pass after this computation step, as well as the leading part of the
backward pass before this step. Once the continuation returns, the gradients (@y.d@ and possibly other
gradients in the closure) should have been correctly updated, and the operator then updates
the gradients of the dependent variables using $\mathrel{+}=$ operations.

\begin{figure}[h!]
\vspace{-4ex}
\begin{multicols}{2}
\begin{lstlisting}[language=Scala,basicstyle=\footnotesize\ttfamily]
// Differentiable real number type.
class NumR0(val x: Double, var d: Double) {
  def +(that: NumR0) = {^\cbox{(k: NumR0=>Unit) =>}^
    val y = new NumR0(this.x + that.x, 0.0); ^\cbox{k(y)}^
    this.d += y.d; that.d += y.d
  }
  def *(that: NumR0) = {^\cbox{(k: NumR0=>Unit) =>}^
    val y = new NumR0(this.x * that.x, 0.0); ^\cbox{k(y)}^
    this.d += that.x * y.d
    that.d += this.x * y.d
  } ...
}
\end{lstlisting}
\columnbreak
\begin{lstlisting}[language=Scala,basicstyle=\footnotesize\ttfamily]]
// Differentiation operator.
def grad(f: NumR0 =>^\cbox{(NumR0=>Unit)=>Unit}^)(x:Double)={
  val z = new NumR0(x, 0.0)
  f(z)((r: NumR0) => r.d = 1.0)
  z.d
}
// Example: 2*x + x*x*x.
val df = grad { x => k =>
  (2*x)^\cbox{(y1=>(\colorbox{white}{x*x})(y2=>(y2\colorbox{white}{*x})(y3=>(y1\colorbox{white}{+}y3)(k))))}^
}
forAll { x =>
  df(x) == 2 + 3*x*x
}
\end{lstlisting}
\end{multicols}
\vspace{-4ex}
\caption{Automatic Differentiation in Scala: reverse-mode AD in continuation-passing style (left),
\texttt{grad} function definition and use case (right). Handling of continuations is highlighted.
Note that \emph{val} and \emph{var} mean immutable and mutable variables respectively in Scala.
Constants are implicitly lifted to \texttt{NumR0}s. Code first appeared in~\cite{wang2018a}.}
\label{fig:CPS1}
\vspace{-2ex}
\end{figure}

However, this implementation is not yet taking care of the generation of delimited continuations.
As a consequence, it is cumbersome to use. Even for our simple running example
$y = 2*x + x*x*x$, we have to explicitly construct delimited
continuations for each step (shaded box in Figure~\ref{fig:CPS1} lower-right).
Fortunately, there exist \emph{delimited control operators} \cite{DBLP:conf/popl/Felleisen88} that enable
programming with delimited continuations in a direct style, without making
continuations explicit. As a next step, we are going to use the @shift@/@reset@ pair of operators \citep{DBLP:conf/lfp/DanvyF90} to simplify our implementation.

\subsection{Implementation using Control Operators}

The @shift@ and @reset@ operators \citep{DBLP:conf/lfp/DanvyF90} work together to capture a partial return path up to a programmer-defined bound:
in our case the remainder of the forward pass.
They are readily available in Scala as a compiler plug-in \citep{DBLP:conf/icfp/RompfMO09},
thus we can simply use them in our @NumR@ implementation.
In Figure~\ref{fig:CPS2}, the keyword @shift@ provides access to a delimited continuation
that reaches up the call chain to the nearest enclosing @reset@.
The Scala compiler transforms all the intermediate code into a continuation,
and passes it to the @shift@ construct as the parameter @k@ \citep{DBLP:conf/icfp/RompfMO09}.
As a result, the implementation of @NumR@ with \texttt{shift}/\texttt{reset} operators is
almost identical to the CPS @NumR0@ implementation in Figure~\ref{fig:CPS1} (modulo the added @shift@).
The implementation also corresponds to formal translation rules we provide in Section~\ref{sec:reverse-formal} and especially Figure~\ref{fig:reverse-formal1}.

The @shift@/@reset@ operators in Scala are tracked by types
annotated in the form of @A@ \at @cps[B]@. Semantically, this means that the @shift@ construct
can be used anywhere @A@-typed values are needed, but it must be
within a @reset@ context of type @B@.
For reverse-mode AD, we expect the
continuation @k@ to be of type @NumR => Unit@, and the body of @shift@ to be of type @Unit@.
\begin{figure}[h!]
\vspace{-5ex}
\begin{multicols}{2}
\begin{lstlisting}[language=Scala,basicstyle=\footnotesize\ttfamily]
// Differentiable number type.
class NumR(val x: Double, var d: Double) {
  def +(that: NumR) = ^\cbox{shift \{(k: NumR=>Unit) =>}^
    val y = new NumR(this.x + that.x, 0.0); ^\cbox{k(y)}^
    this.d += y.d; that.d += y.d
  }
  def *(that: NumR) = ^\cbox{shift \{(k: NumR=>Unit) =>}^
    val y = new NumR(this.x * that.x, 0.0); ^\cbox{k(y)}^
    this.d += that.x * y.d
    that.d += this.x * y.d
  } ...
}
\end{lstlisting}
\columnbreak
\begin{lstlisting}[language=Scala,basicstyle=\footnotesize\ttfamily]
// Differentiation operator.
def grad(f: NumR => NumR^\cbox{@cps[Unit]}^)(x: Double) = {
  val z = new NumR(x, 0.0)
  ^\cbox{reset}^ { f(z).d = 1.0 }
  z.d
}
// Example: 2*x + x*x*x.
val df = grad { x =>
  2*x + x*x*x
}
forAll { x =>
  df(x) == 2 + 3*x*x
}
\end{lstlisting}
\end{multicols}
\vspace{-4ex}
\caption{Automatic Differentiation in Scala: reverse-mode using delimited continuations,
with \texttt{shift}/\texttt{reset} operators (left), \texttt{grad} function definition, and use case (right).
Handling of continuations (shaded boxes) is confined to implementation logic and does not leak into user code.
Constants are implicitly lifted to \texttt{NumR}s. Code first appeared in~\cite{wang2018a}.}
\label{fig:CPS2}
\vspace{-2ex}
\end{figure}

\begin{figure}[h]
\vspace{-1ex}
$$\footnotesize
\ba{rll}
\texttt{Transform}(f) &=& \lambda x. \ \Let{\hat x}{(x, \Ref 0)} \\
                       && \quad \ \ \ \Res{\Let{\hat z}{\At \orb\bra{f}\ \hat x} z' := 1.0}; \\
                       && \quad \ \ \ ! \ x' \\
     \texttt{where } \orb\bra{.}: && \Exp \rightarrow \Exp \ \ \ \text{is defined as below:} \\
     \\
\orb\bra{ c }          &=& c \ \ \text{if} \ c \notin \RR \\
\orb\bra{ c }          &=& \Red{(c, \Ref 0)} \ \ \text{if} \ c \in \RR \\
\orb\bra{ y }                &=& y \\
\orb\bra{ e_1 + e_2 } &=& \Blue{\Let{\hat y_1}{\orb\bra{e_1}}} \\
                       && \Blue{\Let{\hat y_2}{\orb\bra{e_2}}} \\
                       && \Blue{\texttt{shift } k \text{ in } \Let{\hat y}{(y_1 + y_2, \Ref 0)}} \\
                       && \Blue{\quad \quad \quad \quad \quad \At k \ \hat y;} \\
                       && \Blue{\quad \quad \quad \quad \quad y_1' \ \mathrel{+}= \ ! \ y';}\\
                       && \Blue{\quad \quad \quad \quad \quad y_2' \ \mathrel{+}= \ ! \ y'}\\
\orb\bra{ e_1 * e_2 } &=& \Blue{\Let{\hat y_1}{\orb\bra{e_1}}} \\
                       && \Blue{\Let{\hat y_2}{\orb\bra{e_2}}} \\
                       && \Blue{\texttt{shift } k \text{ in } \Let{\hat y}{(y_1 * y_2, \Ref 0)}} \\
                       && \Blue{\quad \quad \quad \quad \quad \At k \ \hat y;} \\
                       && \Blue{\quad \quad \quad \quad \quad y_1' \ \mathrel{+}= \ ! \ y' * y_2;}\\
                       && \Blue{\quad \quad \quad \quad \quad y_2' \ \mathrel{+}= \ ! \ y' * y_1}\\
\orb\bra{ \lambda y. \ e } &=& \lambda y. \ \orb\bra{e} \\
\orb\bra{\At e_1 \ e_2}            &=& \At \orb\bra{e_1} \ \orb\bra{e_2} \\
\orb\bra{\Let{y}{e_1} e_2} &=& \Let{y}{\orb\bra{e_1}} \orb\bra{e_2} \\
\orb\bra{\Fst{e}} &=& \Fst{\orb\bra{e}} \\
\orb\bra{\Snd{e}} &=& \Snd{\orb\bra{e}} \\
\orb\bra{\Ref{e}} &=& \Ref{\orb\bra{e}} \\
\orb\bra{! \ e} &=& ! \ {\orb\bra{e}} \\
\orb\bra{e_1 := e_2} &=& \orb\bra{e_1} := \orb\bra{e_2} \\
\orb\bra{(e_1, e_2)} &=& (\orb\bra{e_1}, \orb\bra{e_2}) \\
\orb\bra{\Left{e}} &=& \Left{\orb\bra{e}} \\
\orb\bra{\Right{e}} &=& \Right{\orb\bra{e}} \\
\orb\bra{\Pat{e}{y_1}{e_1}{y_2}{e_2}}
&=& \Pat{\orb\bra{e}}{y_1}{\orb\bra{e_1}}{y_2}{\orb\bra{e_2}} \\
\ea
$$
\vspace{-3ex}
\footnotesize
\caption{
Transformation of reverse-mode AD with \texttt{shift/reset} and mutable state in the target language
(identical to interpretation except for the handling of environments).
Rules that are different from standard transformation are highlighted in blue.
Note that in arithmetic rules (+ and *), the computations for both forward-pass and backward-pass
are defined in the same rule, with the captured continuation k executed in between.
This programming pattern directly fits the abstract computation flow in Figure~\ref{fig:reverse1} upper-right,
where continuations are triggered in-between forward computations and backward computations.
The transformation is also local.} \label{fig:reverse-formal1}
\vspace{-4ex}
\end{figure}

\begin{figure}[h]
\vspace{-1ex}
$$\footnotesize
\ba{rll}
\texttt{Transform}(f) &=& \underline \lambda x. \ \DLet{\hat x}{\DPair{x}{\DRef 0}} \\
                       && \quad \quad \Dat (\Dat \orb\bra{f} \ \hat x) \ (\WLam z. \ \DLet{\hat z}{z} z' \ \underline{:=}\ 1.0) \underline{;} \\
                       && \quad \ \ \ \DDref{x'} \ \\
          \texttt{where } \orb\bra{.}: && \Exp \rightarrow \Exp \ \ \ \text{is defined as below:} \\
                       \\
\orb\bra{ c }          &=& c \ \ \text{if} \ c \notin \RR \\
\orb\bra{ c }          &=& \Red{\DPair{c}{\DRef 0}} \ \ \text{if} \ c \in \RR \\
\orb\bra{ y }                &=& y \\
\orb\bra{ e_1 + e_2 } &=& \Blue{\SShift{k}{\DLet{\hat y_1}{\orb\bra{e_1}}}} \\
                       && \quad \quad \quad \quad \ \ \ \Blue{\DLet{\hat y_2}{\orb\bra{e_2}}} \\
                       && \quad \quad \quad \quad \ \ \ \Blue{\DLet{\hat y}{\DPair{\DAdd{y_1}{y_2}}{\DRef 0}}} \\
                       && \quad \quad \quad \quad \ \ \ \Blue{\Sat k \ \hat y \underline{;}} \\
                       && \quad \quad \quad \quad \ \ \ \Blue{\DUpdate{y_1'}{\DDref{y'}}\underline{;}} \\
                       && \quad \quad \quad \quad \ \ \ \Blue{\DUpdate{y_2'}{\DDref{y'}}} \\
\orb\bra{ e_1 * e_2 } &=& \Blue{\SShift{k}{\DLet{\hat y_1}{\orb\bra{e_1}}}} \\
                       && \quad \quad \quad \quad \ \ \ \Blue{\DLet{\hat y_2}{\orb\bra{e_2}}} \\
                       && \quad \quad \quad \quad \ \ \ \Blue{\DLet{\hat y}{\DPair{\DMul{y_1}{y_2}}{\DRef 0}}} \\
                       && \quad \quad \quad \quad \ \ \ \Blue{\Sat k \ \hat y\underline{;}} \\
                       && \quad \quad \quad \quad \ \ \ \Blue{\DUpdate{y_1'}{\DMul{\DDref{y'}}{y_2}}\underline{;}} \\
                       && \quad \quad \quad \quad \ \ \ \Blue{\DUpdate{y_2'}{\DMul{\DDref{y'}}{y_1}}} \\
\orb\bra{ \lambda y. \ e } &=& \underline \lambda y. \ \underline \lambda k.
                                    \ \SRes{\Wat k \ \orb\bra{e}} \\
\orb\bra{ \ \At e_1 \ e_2} &=& \SShift{k}{\Dat (\Dat \orb\bra{e_1} \orb\bra{e_2})(\WLam a. \Sat k \ a)} \\
\#  \ \ \ \orb\bra{\Let{y}{e_1} e_2} &=& \SShift{k}{\DLet{y}{\orb\bra{e_1}}{\SRes{\Sat k \ \orb\bra{e_2}}}} \\
\orb\bra{\Fst{e}} &=& \DFst{\orb\bra{e}} \\
\orb\bra{\Snd{e}} &=& \DSnd{\orb\bra{e}} \\
\orb\bra{\Ref{e}} &=& \DRef{\orb\bra{e}} \\
\orb\bra{! \ {e}} &=& \DDref \orb\bra{e} \\
\orb\bra{e_1 := e_2} &=& \orb\bra{e_1} \ \underline{:=} \ \orb\bra{e_2} \\
\orb\bra{(e_1, e_2)} &=& \DPair{\orb\bra{e_1}}{\orb\bra{e_2}} \\
\orb\bra{\Left{e}}  &=& \DLeft{\orb\bra{e}}  \\
\orb\bra{\Right{e}} &=& \DRight{\orb\bra{e}} \\
\orb\bra{\Pat{e}{y_1}{e_1}{y_2}{e_2}} &=&
     \SShift{k}{\WLet{k_1}{\WLam a. \ \Sat k \ a}\\
  && \DPat{\orb\bra{e}}{y_1}{\SRes{\Wat k_1 \ \orb\bra{e_1}}}
                       {y_2}{\SRes{\Wat k_1 \ \orb\bra{e_2}}}} \\
\textbf{normalization rules for wavy underline}: \\
\WLam y. \ \Wat e \ y &\rightarrow& e \\
\WLet{y}{y_1} e       &\rightarrow& e[y \leftarrow y_1]
\ea
$$
\vspace{-4ex}
\caption{
Transformation of reverse-mode AD with \texttt{shift/reset}
in the meta-language.
Rules that are different from standard transformation are labeled in blue.
The standard rules are adapted from \citet{DBLP:journals/mscs/DanvyF92},
and the \# symbol denotes rules that are simplified due to Barendregt's variable convention.
We also adapted the overline/underline notation from \citet{DBLP:journals/mscs/DanvyF92},
such that the overline denotes static/meta-language constructs, and the underline
denotes dynamic/target-language constructs. Departing slightly from \citet{DBLP:journals/mscs/DanvyF92},
we introduce another \textbf{wavy underline notation} to implement proper tail calls.
Wavy underline denotes target-language terms just as normal underline, but wavy terms
will be normalized with respect to the contraction rules while the target
expression is built up (lower part of this figure).
Note that wavy underline normalization does only renaming, not full substitution.
}\label{fig:reverse-formal2}
\vspace{-3ex}
\end{figure}

\begin{figure}[h]
\vspace{-3ex}
$$\footnotesize
\ba{rll}
\texttt{Transform}(f) &=& \underline \lambda x. \ \DLet{\hat x}{\DPair{x}{\DRef 0}} \\
                       && \quad \ \ \ \Sat \orb\bra{f} \ (\overline \lambda m. \ \Dat ( \Dat m \ \hat x) \ (\WLam z. \ \DLet{\hat z}{z} z' \ \underline{:=} \ 1.0))\underline{;} \\
                       && \quad \ \ \ \DDref{x'} \ \\
          \texttt{where } \orb\bra{.}: && \Exp \rightarrow \Exp \ \ \ \text{is defined as below:} \\
                       \\
\orb\bra{ c }         &=& \overline \lambda \kappa.\ \Sat \kappa \ c \ \ \text{if} \ c \notin \RR \\
\orb\bra{ c }         &=& \overline \lambda \kappa.\ \Sat \kappa \ \Red{\DPair{c}{\DRef 0}} \ \ \text{if} \ c \in \RR \\
\orb\bra{ y }         &=& \overline \lambda \kappa.\ \Sat \kappa \ y \\
\orb\bra{ e_1 + e_2 } &=& \overline \lambda \kappa. \
                         \Blue{\Sat \orb\bra{e_1}(\overline \lambda p_1. \ \Sat \orb\bra{e_2}(\overline \lambda p_2.} \ \ \#\# \\
                      && \quad \quad \Blue{\DLet{\hat y_1}{p_1}} \ \Blue{\DLet{\hat y_2}{p_2}} \\
                      && \quad \quad \Blue{\DLet{\hat y}{\DPair{\DAdd{y_1}{y_2}}{\DRef 0}}} \\
                      && \quad \quad \Blue{\Sat \kappa \ \hat y\underline{;}}\\
                      && \quad \quad \Blue{\DUpdate{y_1'}{\DDref y'}\underline{;}}\\
                      && \quad \quad \Blue{\DUpdate{y_2'}{\DDref y'}))}\\
\orb\bra{ e_1 * e_2 } &=& \overline \lambda \kappa. \
                         \Blue{\Sat \orb\bra{e_1}(\overline \lambda p_1. \ \Sat \orb\bra{e_2}(\overline \lambda p_2.} \ \ \#\# \\
                      && \quad \quad \Blue{\DLet{\hat y_1}{p_1}} \ \Blue{\DLet{\hat y_2}{p_2}} \\
                      && \quad \quad \Blue{\DLet{\hat y}{\DPair{\DMul{y_1}{y_2}}{\DRef 0}}} \\
                      && \quad \quad \Blue{\Sat \kappa \ \hat y\underline{;}} \\
                      && \quad \quad \Blue{\DUpdate{y_1'}{\DMul{\DDref y'}{y_2}}\underline{;}} \\
                      && \quad \quad \Blue{\DUpdate{y_2'}{\DMul{\DDref y'}{y_1}}))} \\
\orb\bra{ \lambda y. \ e } &=& \overline \lambda \kappa. \ \Sat \kappa \ (\underline \lambda y. \ \underline \lambda k.
                                  \ \Sat \orb\bra{e} (\SLam m. \ \Wat k \ m)) \\
\orb\bra{ \ \At e_1 \ e_2}
&=& \overline \lambda \kappa. \ \Sat \orb\bra{e_1}(\overline \lambda m. \ \Sat \orb\bra{e_2}(\overline \lambda n. \
    \Dat (\ \Dat m \ n) \ (\WLam a. \ \Sat \kappa \ a))) \\
\#  \ \ \ \orb\bra{\Let{y}{e_1} e_2}
&=& \overline \lambda \kappa. \ \Sat \orb\bra{e_1}(\overline \lambda y_1. \ \DLet{y}{y_1} \Sat \orb\bra{e_2} \ \kappa) \\
\orb\bra{\Fst{e}} &=& \overline \lambda \kappa. \ \Sat \orb\bra{e}(\overline \lambda y. \ \Sat \kappa \ (\DFst{y})) \\
\orb\bra{\Snd{e}} &=& \overline \lambda \kappa. \ \Sat \orb\bra{e}(\overline \lambda y. \ \Sat \kappa \ (\DSnd{y})) \\
\orb\bra{\Ref{e}} &=& \overline \lambda \kappa. \ \Sat \orb\bra{e}(\overline \lambda y. \ \Sat \kappa \ (\DRef{y})) \\
\orb\bra{! \ {e}} &=& \overline \lambda \kappa. \ \Sat \orb\bra{e}(\overline \lambda y. \ \Sat \kappa \ (\DDref{y})) \\
\orb\bra{e_1 := e_2} &=& \overline \lambda \kappa. \ \Sat \orb\bra{e_1}(\overline \lambda y_1. \ \Sat \orb\bra{e_2}
   (\overline \lambda y_2. \ \Sat \kappa \ ( y_1 \ \underline{:=} \ y_2))) \\
\orb\bra{(e_1, e_2)} &=& \overline \lambda \kappa. \ \Sat \orb\bra{e_1}(\overline \lambda y_1. \ \Sat \orb\bra{e_2}
   (\overline \lambda y_2. \ \Sat \kappa \ (\DPair{y_1}{y_2}))) \\
\orb\bra{\Left{e}}  &=& \overline \lambda \kappa. \ \Sat \orb\bra{e}(\overline \lambda y. \ \Sat \kappa \ (\DLeft{y})) \\
\orb\bra{\Right{e}} &=& \overline \lambda \kappa. \ \Sat \orb\bra{e}(\overline \lambda y. \ \Sat \kappa \ (\DRight{y})) \\
\orb\bra{\Pat{e}{y_1}{e_1}{y_2}{e_2}}
&=& \overline \lambda \kappa. \ \WLet{k}{\WLam a. \ \Sat \kappa \ a} \Sat \orb\bra{e}(\overline \lambda v. \
                                \DPatL{v}{y_1}{\Sat \orb\bra{e_1}(\SLam m. \ \Wat k \ m)}
                                         {y_2}{\Sat \orb\bra{e_2}(\SLam n. \ \Wat k \ n)}) \\
\ea
$$
\vspace{-3ex}
\caption{
  Transformation of source language for reverse-mode AD in CPS (meta-language does not contain \texttt{shift/reset}).
  Rules that are different from standard CPS transformation are highlighted by color.
  Note that in the plus rule and the multiplication rule (labeled by \#\#),
  we avoided using variable sugaring in $\overline \lambda p_1$ and $\overline \lambda p_2$
  so that we can introduce a dynamic \texttt{let} binding for them. The dynamic \texttt{let} bindings are necessary
  to preserve sharing, evaluation order, and asymptotic complexity, since the right-hand sides of them are accessed multiple times
  via $y_1, y_1', y_2$, and $y_2'$.
}\label{fig:reverse-formal3}
\vspace{-4ex}
\end{figure}

\subsection{Reverse-Mode AD for Lambda Calculus}
\label{sec:reverse-formal}

We now formalize reverse-mode AD as a transformation based on the same lambda calculus as used for the forward mode (Figure ~\ref{fig:formal}).
The straightforward first step is to
make use of @shift@/@reset@ control operators in the target language to capture
continuations delimited at the end of AD computation. We provide formal rules for
this transformation in Figure~\ref{fig:reverse-formal1}, matching the Scala implementation in Figure~\ref{fig:CPS2}.
Note that the arrow of the new differentiation operator $\orb\bra{.}$ indicates reverse-mode.
Similar to the forward differentiation operator, the differentiation is still with respect to
a fixed $x$, but the $\orb\bra{.}$ operator never encounters the special $x$, since
we only transform the abstraction $\lambda x. e$, assuming Barendregt's variable convention.

Making use of @shift@/@reset@ control operators in the target language, the formal rules
in Figure~\ref{fig:reverse-formal1} precisely capture the idea of the abstract nested computation flow
in Figure~\ref{fig:reverse1} and the Scala implementation in Figure~\ref{fig:CPS2}. However, what if we want to use a target language that does not
provide @shift@/@reset@ operators? This can be achieved by moving the uses of @shift@/@reset@
into the meta-language \cite{DBLP:journals/mscs/DanvyF92} (so that they are used at the time of translation), and generating
target terms in explicit CPS (without @shift@/@reset@). We provide formal rules for this transformation
in Figure~\ref{fig:reverse-formal2}.
The \emph{result} of this translation matches the
Scala implementation in Figure~\ref{fig:CPS1}.
Note that in this and following figures, we
use overline/underline notations (adapted from \citet{DBLP:journals/mscs/DanvyF92}) to mark
static/metalanguage constructs (overline), and dynamic/target language constructs (underline).
We also introduce a wavy underline notation for handling proper tail calls, with special
reduction-upon-construction logic (Figure~\ref{fig:reverse-formal2} lower).
Note that the wavy underline notation for ``let'' means that @let@ bindings should be
removed if and only if the right-hand side of the @let@ binding is just a variable/symbol,
so this wavy underline normalization performs only renaming, not full substitution.
This rule is not strictly necessary for properly tail-recursive calls, but it removes
unnecessary symbol bindings for the $\texttt{case}$ expression in abstraction
(supporting implementation of this transformation in Scala with examples is available online\textsuperscript{\ref{github:supplemental}}).

It is of course also possible to express the CPS transformation without @shift@/@reset@
entirely by switching the meta-language code to CPS. This can be achieved formally
by applying the same transformation as above to the meta-language translation code.
The result is that occurrences of $\text{shift/reset}$ are fully erased from the right-hand sides of the
translation (Figure~\ref{fig:reverse-formal3}).
A complete version of the formal presentation
with standard interpretations/transformations and examples of loops and recursions is
available online\textsuperscript{\ref{github:supplemental}}.

\subsection{Relation to Previous Functional Approaches}

It is important to note at this point that our description of reverse-mode AD may appear similar to
\citet{DBLP:journals/toplas/PearlmutterS08}. However, there are substantial differences
despite the similarities and shared goals.
The implementation proposed by Pearlmutter and Siskind returns a pair of a value and
a backpropagator: $x \mapsto (v, dy/dv \mapsto dy/dx)$ for backward propagation.
Doing this correctly requires a non-local program transformation, as noted in that paper.
Further tweaks are required if a lambda uses variables from an outer scope: there must be some
mechanism that allows backpropagation for captured variables, not just the function inputs.

In contrast to \citet{DBLP:journals/toplas/PearlmutterS08},
using delimited continuations with @shift@/@reset@ operators enables
reverse-mode AD with only local transformations. Any underlying non-local
transformations are implicitly resolved by @shift@ and @reset@.
Beyond this, it is also worth
noting that our method can allocate all closures and mutable variables on the stack, i.e, we
never need to return closures that escape their allocation scope.
The proposed implementation is also extremely concise, to the point that it can serve as a
\emph{specification} of reverse-mode AD and can be used to teach AD to students.

\subsection{Relation to Tape-Based Approaches}

From our implementation using delimited continuations, we can
derive a classic tape-based formulation of reverse-mode AD.
We first realize that conceptually, our use of delimited
continuations builds an implicit representation of a tape-like
structure on the call stack instead of representing it as
an explicit data structure on the heap. We can map this
implicit structure back to the heap, by accumulating
the gradient-update code that follows the invocations
of @k@ into closures, and storing their composition
in a global mutable variable, which is used to explicitly
invoke the backward pass. After this change, all invocations
of continuations become tail calls, and hence delimited
continuations or control operators are no longer necessary.
The downside of this approach is the potentially costly
management of heap-allocated closures,
and, crucially, a less straightforward mapping to staged or \emph{define-then-run} AD implementations that reify computation graphs, which falls out very naturally for CPS-based formulations (see Section~\ref{sec:lms}).

We show a Scala implementation in
Figure~\ref{fig:relay}, noting that a similar implementation
has been proposed by \citet{DBLP:journals/corr/abs-1810-00952}
in their framework Relay. It is easy to see that this
implementation can be defunctionalized \citep{DBLP:journals/lisp/Reynolds98a, DBLP:conf/ppdp/DanvyN01} to obtain a classic
tape-based AD formulation, and thus can be seen as a
refunctionalized version \citep{DBLP:journals/scp/DanvyM09}
of such a classic tape datastructure.

\begin{figure}[h!]
\vspace{-5ex}
\begin{multicols}{2}
\begin{lstlisting}[language=Scala,basicstyle=\footnotesize\ttfamily]
// Refunctionalized tape.
var tape = (u: Unit) => ()
// Differentiable number type.
class NumB(val x: Double, var d: Double) {
  def +(that: NumB) = {
    val y = new NumB(this.x + that.x, 0.0)
    tape = ((x:Unit)=> this.d += y.d) andThen tape
    tape = ((x:Unit)=> that.d += y.d) andThen tape
    y
  }
  def *(that: NumB) = {
    val y = new NumB(this.x * that.x, 0.0)
    tape = ((x:Unit)=> this.d += that.x*y.d) andThen tape
    tape = ((x:Unit)=> that.d += this.x*y.d) andThen tape
    y
  } ...
}
\end{lstlisting}
\columnbreak
\begin{lstlisting}[language=Scala,basicstyle=\footnotesize\ttfamily]
// Differentiation operator.
def grad(f: NumB => NumB)(x: Double) = {
  val z = new NumB(x, 0.0)
  f(z).d = 1.0
  tape()
  z.d
}

// Example: 2*x + x*x*x.
val df = grad { x =>
  2*x + x*x*x
}

forAll { x =>
  df(x) == 2 + 3*x*x
}
\end{lstlisting}
\end{multicols}
\vspace{-4ex}
\caption{Reverse-mode Automatic Differentiation with a global refunctionalized tape.
The andThen infix operator is for function composition in Scala.
Since new additions are composed before the old tape, calling $\texttt{tape()}$ will play the tape in reverse order of insertion.}
\label{fig:relay}
\vspace{-2ex}
\end{figure}

\subsection{Purely Functional Implementation}\label{sec:purelyfunc}

Since our presentation makes central use of the mutable state, an interesting question is whether a purely
functional formulation is also possible. For example, since the continuation @k@ takes a new @NumR@,
updates its gradient, and returns @Unit@, why not simply let @k@ return the new gradient and
avoid mutation? The type of @k@ would change to @Double => Double@ accordingly.
Unfortunately, this simple change is not enough, because the continuation @k@ may update the
gradients of \emph{more than one} @NumR@s.
If earlier @NumR@s are also involved in the computations in @k@, then @k@ needs to update their gradients too,
but returning just a @Double@ without side-effects cannot achieve that.
Thus, a pure functional implementation is easy to achieve for straight-line programs \citep{DBLP:journals/pacmpl/Elliott18},
but not for ones with complex control flow and especially nested lambdas.

Based on this observation, we can build a purely functional implementation by adding a layer of indirection.
Each @NumR@ is assigned a unique id, and we change the type of continuations to @NumR => Map[Id, Double]@,
returning an immutable map from @NumR@ ids to their calculated gradient updates. In essence, this model
uses a reified functional store for gradient updates instead of storing the gradients directly in the
Scala heap. Since there is no conceptual simplification, we prefer the model based on direct mutation
for our presentation.

\subsection{Nested Invocations For Higher-Order Gradients}\label{sec:highorderreverse}

Just like with forward-mode AD in Section~\ref{sec:highorderforward},
we are interested in extending the reverse-mode AD implementation to
support nested invocations of the @grad@ operator.
The way to achieve this nesting of reverse-mode
AD within reverse-mode AD (i.e., reverse-of-reverse) is to use multiple
levels of continuations, and their corresponding higher-order control
operators such as @shift2@~\cite{DBLP:conf/lfp/DanvyF90}.
Unfortunately, we cannot directly implement this in Scala,
since the Scala compiler only provides a single CPS transform layer.
However, we can manually embed a @shift@/@reset@ layer within another
@shift@/@reset@ via explicit CPS to create a similar functionality as @shift2@ (Figure~\ref{fig:shift2}).

\begin{figure}[h!]
\vspace{-5ex}
\begin{multicols}{2}
\begin{lstlisting}[language=Scala,basicstyle=\footnotesize\ttfamily]
// Type of contexts (explicit or implicit CPS).
type Ctx  = ((Unit => Unit) => Unit)
type Ctxi = Unit @cps[Unit]
// Definition of (restricted) shift2.
def shift2(body:(NumRR=>Ctxi)=>Ctxi): NumRR @cps[Ctx] =
  shift { k: (NumRR => Ctx) => k2: (Unit => Unit) =>
    def kk(y: NumRR) = shift((k3: Unit=>Unit) => k(y)(k3))
    reset { body(kk); k2() }
  }
// Differentiable number class.
class NumRR(val x: NumR, var d: NumR) {
  def * (that: NumRR) = shift2 {k: (NumRR => Ctxi) =>
    val y = new NumRR(x * that.x, new NumR(0.0,0.0))
    k(y)
    this.d = this.d + y.d * that.x
    that.d = that.d + y.d * this.x
  }
  ...
}
\end{lstlisting}
\columnbreak
\begin{lstlisting}[language=Scala,basicstyle=\footnotesize\ttfamily]
// Differentiation operator.
def gradRR(f: NumRR => NumRR @cps[Ctx])(x: NumR) =
  shift { (k: NumR => Unit) =>
    val z = new NumRR(x, new NumR(0.0,0.0))
    val ff = reset {
      f(z).d = new NumR(1.0,0.0)
      (k2: Unit => Unit) => k2()
    }
    ff((u: Unit) => k(z.d))
  }

// Example: 2*x + x*x*x.
val df = grad { gradRR { x =>
  2*x + x*x*x
}}
forAll { x =>
  df(x) == 6*x
}
\end{lstlisting}
\end{multicols}
\vspace{-4ex}
\caption{Second order reverse-of-reverse AD via explicit CPS in \texttt{shift/reset}.}
\label{fig:shift2}
\vspace{-2ex}
\end{figure}

Another way to realize higher-order gradients is to nest forward-mode AD in
reverse-mode AD (i.e. compute the first-order gradient via reverse-mode AD, and
higher order gradient via forward-mode AD). This approach is practically
efficient if the higher-order gradient is for functions $\RR^m \rightarrow \RR^n$,
where $n$ is relatively large compared to $m$.
This "forward-of-reverse" combination can efficiently compute Hessians as the Jacobian of gradients
\cite{DBLP:journals/corr/BaydinPR18},
and Hessian-vector products in a single forward-of-reverse pass \cite{christianson1992automatic}.

\section{Reifying Computation Graphs via Multi-Stage Programming}\label{sec:lms}

CPS conversion puts reverse-mode AD on a firm basis, rooted in programming language concepts.
Extending the @Num@ type to tensors and relaying tensor operations to high performant libraries
provides all the necessary machinery for a deep learning framework
in the expressive PyTorch-style that performs gradient computation
as part of the normal program execution (``define-by-run'').
\vspace{-1ex}
\paragraph{From PyTorch-style to TensorFlow-style}
\ However, TensorFlow-style frameworks have traditionally been more performant
than define-by-run ones, by constructing a restricted
dataflow model \emph{before} executing gradient computation,
which offers a larger optimization surface on the tensor IR level (``define-then-run'').
Can we also realize a TensorFlow-style framework, but with a richer and more standard IR language,
better supporting native control flow and recursion?

\vspace{-1ex}
\paragraph{TensorFlow-style via Multi-Stage Programming}
\ This question can be naturally addressed by leveraging the idea found in the formal model of moving the use of shift/rest into the metalanguage to generate code in CPS (Section~\ref{sec:reverse}). We use \emph{multi-stage programming} (staging) as a practical way to realize
the overline/underline distinction found in the formal model.
Modern tools such as LMS (Lightweight Modular Staging)~\citep{DBLP:conf/gpce/RompfO10}
blend normal program execution with IR construction.
In LMS, a type constructor @Rep[T]@ is used to mark staged expressions. That is to say,
all @Rep[T]@-typed variables (whether directly labeled or type-inferred)
will trigger LMS-based IR construction. Through type inference and advanced operator overloading,
normal syntax can be used to stage built-in control-flow constructs such as @if@, @for@, and @while@.
We can relate staging via LMS to the formal rules in Figure~\ref{fig:reverse-formal2}, though
the different stages are determined by types in LMS \cite{DBLP:conf/birthday/Rompf16}.

To show the flavor of LMS as well as how to make use of LMS in our reverse-mode AD
to reify computation graphs (LMS-based IR), let us walk through our running example again:
$y = 2 * x + x * x * x$, where we simply focus on first-order reverse-mode AD such that $x$ is of type @NumR@.
To stage our running example,
the most important change is the type signature of the @NumR@ class:
\begin{lstlisting}[language=Scala,basicstyle=\footnotesize\ttfamily]
class NumR(val x: Rep[Double], val d: Rep[Var[Double]]) {...}
\end{lstlisting}
Here, the @Rep[T]@ type of @x@ and @d@ states that all handling of @x@ and @d@ will construct nodes in LMS-IR.
The @Rep[Var[Double]]@ maps to staged mutable reference (such as type @double&@ in C++),
which allows us to accumulate gradients @d@ by reference.
Note that our presentation is isomorphic to staging @NumR@ as in @Rep[NumR]@,
since both fields of @NumR@ are already staged.
However, staging only the fields of @NumR@ gives us a more concise generated code.

There are no fundamental challenges with staging our reverse-mode AD in CPS using LMS, as it is a well-known insight that
multi-stage programs that use continuations at generation time can generate code in CPS
\cite{LFP-1992-Bondorf, DBLP:journals/mscs/DanvyF92} (relating to formal rules in Figure~\ref{fig:reverse-formal2}).
LMS can also be set up to generate low-level, efficient code in C++ and CUDA.
This enables a TensorFlow-style framework with rich analysis and optimization
opportunities, much like an aggressive whole-program compiler.

The apparent downsides of TensorFlow-style systems, however, are the rather
clunky user programming model offered by current frameworks, the absence of
sophisticated control flow constructs, and the inability to use standard
debugging facilities. However, our system largely avoids the downsides of
current static frameworks thanks to staging (in particular, the LMS framework).
Of course, TensorFlow can also be viewed as a staged programming model, but the
staged language is a restricted dataflow language. On the other hand, LMS provides
a rich staged language that includes subroutines, recursion, and more.

We show below how CPS code generation is supported in a natural form, in straight-line code, branches,
loops, and recursion. Note that our setup is mostly similar to Figure~\ref{fig:reverse-formal2},
where only the metalanguage has @shift@/@reset@, but not identical, since stages are controlled by types
(more redexes can be simplified). Also, we will
refer to generic types (@A@, @B@, and @C@) for control flow constructs in the following part of this section to
illuminate the abstraction of branches, loops, and recursion.

\vspace{-1ex}
\subsection{Staging Reverse-Mode AD: Straight-Line Code}

We begin by investigating how to stage and perform AD on straight-line programs
(i.e., those without loops, branches, or recursion).
Let us start with a very simple straight-line program.

\begin{lstlisting}[language=Scala,basicstyle=\footnotesize\ttfamily]
def snippet(in: Rep[Double]): Rep[Double] = grad(x => x * x)(in)
\end{lstlisting}
We show the code after reducing arithmetic operations, @grad@ function, and @shift@/@reset@ control operators (left),
and the generated pseudo-LMS-IR (middle). Note that in this example,
since the @NumR@ class itself is not @Rep[T]@-typed (both fields of @NumR@ are),
fields of @NumR@ will trigger IR-construction for code generation, but all @NumR@
object construction and field accesses will be staged away.
The IR can be used to generate C++ code (with optimizations
including dead code elimination, constant folding, etc., shown on the right):\\
\begin{minipage}[t]{0.45\textwidth}
\begin{lstlisting}[language=Scala,basicstyle=\footnotesize\ttfamily]
def snippet(in: Rep[Double]): Rep[Double] = {
  val z = new NumR(in, 0.0)
  val y = new NumR(z.x * z.x, 0.0)
  y.d = 1.0
  z.d += y.d * z.x
  z.d += y.d * z.x
  z.d
}
\end{lstlisting}
\end{minipage}
\begin{minipage}[t]{0.28\textwidth}
\begin{lstlisting}[language=Scala,basicstyle=\footnotesize\ttfamily]
def snippet(in) = {
  d0 = ref 0
  v1 = in * in; d1 = ref 0
  d1 := 1.0
  d0 += ! d1 * in
  d0 += ! d1 * in
  ! d0
}
\end{lstlisting}
\end{minipage}
\begin{minipage}[t]{0.18\textwidth}
\begin{lstlisting}[language=Scala,basicstyle=\footnotesize\ttfamily]
double snippet(in: double) =
  return 2 * in;
\end{lstlisting}
\end{minipage}

\subsection{Staging Reverse-Mode AD: Conditionals}\label{sec:IF}

The conditionals are closely related to the $\texttt{case}$ rule in Figure~\ref{fig:reverse-formal2}
(e.g., @reset@ in both branches).
We define a syntactically different @IF@ operator that takes
a @Rep[Boolean]@ condition and two @(=> Rep[A]@ \at @cps[Rep[B]])@ typed parameters
for the @then@- and @else@-branches.
In Scala, @=> T@ typed parameters are \emph{passed by name}, so that the parameters are evaluated each time they are used.
Following the $\texttt{case}$ rule, the @IF@ function accesses the delimited continuation @k@ via @shift@,
lifts @k@ to a dynamic function @k1@ (to avoid code duplication), and
invokes @k1@ with both the @then@-branch and the @else@-branch argument.
In LMS, the @fun@ function lifts static functions to dynamic ones~\cite{DBLP:conf/birthday/Rompf16}:
\begin{lstlisting}[language=Scala,basicstyle=\footnotesize\ttfamily]
def fun(f: Rep[A] => Rep[B]): Rep[A => B]
\end{lstlisting}
We use overline/underline to mark function names in function applications and definitions,
and control flow constructs to indicate their stages (meta-language vs target language).
This is similar to the formal rules in Section~\ref{sec:reverse}, but added here purely for reasons of readability.
\begin{center}
\begin{lstlisting}[language=Scala,basicstyle=\footnotesize\ttfamily,escapechar=\#]
def #$\overline{\text{IF}}$#(c: Rep[Boolean])(a: => Rep[A] @cps[Rep[B]])(b: => Rep[A] @cps[Rep[B]]): Rep[A] @cps[Rep[B]] =
  #$\overline{\text{shift}}$# { k: (Rep[A] => Rep[B]) =>
    // Emit k1 as a dynamic function to avoid code duplication.
    val k1 = #$\overline{\text{fun}}$# (k)
    // Emit conditional, with each branch enclosed by a reset.
    #\underline{if}# (c) #$\overline{\text{reset}}$#(#\underline{k1}#(a)) #\underline{else}# #$\overline{\text{reset}}$#(#\underline{k1}#(b))
  }
\end{lstlisting}
\end{center}
Below is an example using the @IF@ construct. For readability, we
only selectively label \emph{some} names with overline/underline to highlight stages of \emph{some} constructs.
\begin{lstlisting}[language=Scala,basicstyle=\footnotesize\ttfamily]
def snippet(in: Rep[Double]): Rep[Double] = grad(x => IF(x.x > 0.0){ -1.0*x*x }{ x*x })(in)
\end{lstlisting}
We show the code after we reduce @grad@, @fun@, and some @shift@/@reset@ (upper left),
the pseudo-IR (right),
and the generated C++ code with optimizations including inlining and hoisting (lower left):
\\
\begin{minipage}[t]{0.65\textwidth}
\begin{lstlisting}[language=Scala,basicstyle=\footnotesize\ttfamily,escapechar=\#]
def snippet(in: Rep[Double]): Rep[Double] = {
  val z = new NumR(in, 0.0)
  val k1: Rep[NumR => Unit] = (t => t.d = 1.0)
  // elide process of reset block (similar to straight-line program)
  #\underline{if}# (z.x > 0.0) #$\overline{\text{reset}}$# { #\underline{k1}# (-1.0 * z * z) }
  // elide process of reset block (similar to straight-line program)
  #\underline{else}# #$\overline{\text{reset}}$# { #\underline{k1}# (z * z) }
  z.d
}

double Snippet(double in) {
  auto k = [&](double x, double& d) { d = 1.0; };
  double d = 0.0;
  if (in > 0.0) { k(-in * in, d); return -2.0 * in * d; }
  else { k(in * in, d); return 2.0 * in * d; }
}
\end{lstlisting}
\end{minipage}
\begin{minipage}[t]{0.32\textwidth}
\begin{lstlisting}[language=Scala,basicstyle=\footnotesize\ttfamily]
def snippet(in) = {
  d0 = ref 0
  k = (x, d) => d := 1.0
  if (in > 0.0) {
    v1 = - in * in; d1 = ref 0
    k(v1, d1)
    d0 += ! d1 * (- in)
    d0 += ! d1 * (- in)
    ! d0
  } else {
    v1 = in * in; d1 = ref 0
    k(v1, d1)
    d0 += ! d1 * in
    d0 += ! d1 * in
    ! d0
  }
}
\end{lstlisting}
\end{minipage}
\vspace{-2ex}

\subsection{Staging Reverse-Mode AD: Loops}

Differentiable loop constructs are important for deep learning, for example in recurrent neural networks.
By the rules of CPS transformation, loops need to be transformed into tail-recursive functions.
A loop construct consists of an initial value @Rep[A]@, a loop guard, and a loop body of type @Rep[A] => Rep[A]@ \at @cps[Rep[B]]@ as parameters.
The loop guard can be either @Rep[A] => Rep[Boolean]@, like a @while@ construct, or simply a @Rep[Int]@, like a @for@ construct.
The actual loop logic can be described as follows: if the loop guard is true,
recursively call the loop after invoking the loop body; else call the continuation.
The @WHILE@ construct is defined below, mimicking the standard @while@ loop.
\begin{center}
\begin{lstlisting}[language=Scala,basicstyle=\footnotesize\ttfamily,escapechar=\#]
def #$\overline{\text{WHILE}}$#(init: Rep[A])(c: Rep[A] => Rep[Boolean])(b: Rep[A] => Rep[A] @cps[Rep[B]]): Rep[A] @cps[Rep[B]] =
  #$\overline{\text{shift}}$# { k: (Rep[A] => Rep[B]) =>
    // tail recursive function implementing loop semantics.
    def #\underline{loop}#: Rep[A => B] = #$\overline{\text{fun}}$# { (x: Rep[A]) =>
      #\underline{if}# (#$\overline{\text{c}}$#(x)) #$\overline{\text{reset}}$#(#\underline{loop}#(#$\overline{\text{b}}$#(x))) #\underline{else}# #$\overline{\text{reset}}$#(#$\overline{\text{k}}$#(x))
    }
    #\underline{loop}#(init)
  }
\end{lstlisting}
\end{center}
Below is an example using the @WHILE@ construct:
\begin{lstlisting}[language=Scala,basicstyle=\footnotesize\ttfamily]
def snippet(in: Rep[Double]): Rep[Double] = grad(x => WHILE(x)(t => t.x > 1.0)(t => t * 0.5))(in)
\end{lstlisting}
We show the code after we reduce @grad@, @fun@, some @shift@/@reset@, and application of @b@, @c@, and @k@ (left),
the pseudo-IR (middle), and the generated C++ code (right):
\\
\begin{minipage}[t]{0.40\textwidth}
\begin{lstlisting}[language=Scala,basicstyle=\footnotesize\ttfamily,escapechar=\#]
def snippet(in: Rep[Double]): Rep[Double]= {
  val z = new NumR(in, 0.0)
  def #\underline{loop}#: Rep[NumR => Unit] = (t =>
    // elide process of reset block
    // similar to straight-line program
    #\underline{if}# (t.x > 1.0) #$\overline{\text{reset}}$#(#\underline{loop}#(t * 0.5))
    #\underline{else}# t.d = 1.0
  )
  #\underline{loop}#(z)
  z.d
}
\end{lstlisting}
\end{minipage}
\begin{minipage}[t]{0.25\textwidth}
\begin{lstlisting}[language=Scala,basicstyle=\footnotesize\ttfamily]
def snippet(in) = {
  d0 = ref 0
  def loop = (x, d) => {
    if (x > 1.0) {
      v1 = x * 0.5; d1 = ref 0
      loop(v1, d1)
      d += 0.5 * ! d1
    } else d := 1.0
  }
  loop(in, d0)
  ! d0
}
\end{lstlisting}
\end{minipage}
\begin{minipage}[t]{0.32\textwidth}
\begin{lstlisting}[language=C,basicstyle=\footnotesize\ttfamily]
double Snippet(double in) {
  double d = 0.0;
  auto loop = [&](double x, double& d) {
    if (x > 1.0) {
      double d1 = 0.0;
      loop(0.5 * x, d1);
      d += 0.5 * d1;
    } else d = 1.0;
  };
  loop(in, d);
  return d;
}
\end{lstlisting}
\end{minipage}

We can also relate our @WHILE@ definition to the formal rules.
Though we did not include
an explicit $\texttt{letrec}$ rule in Figure~\ref{fig:reverse-formal2},
Figure~\ref{fig:formal} explains how $\texttt{letrec}$ can be derived from
$\texttt{let}$, $\lambda$, $\At$, and $\texttt{case}$ constructs, the rules of which are identical
to the standard transformations in Figure~\ref{fig:reverse-formal2}. Thus, the $\texttt{letrec}$
rule is also the same as in the standard transformation \citep{DBLP:journals/mscs/DanvyF92}
(modulo wavy underline notation), which we recap here:
\begin{footnotesize}
$$\# \ \ \ \orb\bra{\LetRec{f}{\lambda x. e_1} e_2} = \SShift{k}{\DLetRec{f}{\DLam x. \DLam k_1. \SRes{\Wat k_1 \ \orb\bra{e_1}}} \SRes{\Sat k \ \orb\bra{e_2}}}$$
\end{footnotesize}
Note that the rule is simplified due to Barendregt's variable convention (no variable substitution needed).
However, specific constraints of loops guarantee that the recursive calls always appear in tail positions,
and the recursive function should only be applied once, in the original location of the loop.
That means the abstraction of continuation ($k_1$ in the rule) can be optimized away, just like the
contification optimization for closures. The formal rule should now be adapted (note that
the transformation rule of $\texttt{apply}$ for $f$ should also adapt accordingly) to:
\begin{footnotesize}
$$\# \ \ \ \orb\bra{\LetRec{f}{\lambda x. e} \At f y} = \SShift{k}{\DLetRec{f}{\DLam x. \SRes{\Sat k \ \orb\bra{e}}} \Dat f \ y}$$
\end{footnotesize}
The abstract body $e$ in the above rule is most likely a conditional construct that needs to be translated to @IF@.
Our @WHILE@ definition can be derived after normalization of @IF@ (containing @shift@) with its surrounding @reset@ context.

\subsection{Staging Reverse-Mode AD: Functions \& Recursion}\label{sec:tree_abs}

As a true differentiable programming framework, we aim to handle general forms of recursion.
This is useful in deep learning: one application is processing tree-structured data,
such as sentence parse trees (see Section~\ref{sec:eval}).
We have already seen in Section~\ref{sec:IF} how we can use @fun@ to generate staged functions in LMS, but how do we make those differentiable? The answer is simply to provide an equivalent of @fun@ that generates a staged function in CPS (@FUN@ below):
\begin{center}
\begin{lstlisting}[language=Scala,basicstyle=\footnotesize\ttfamily,escapechar=\#]
def #$\overline{\text{FUN}}$#($\text{\Blue{f}}$: Rep[A] => Rep[B] @cps[Rep[C]]) = (y: Rep[A]) => #$\overline{\text{shift}}$# {k: (Rep[B] => Rep[C]) =>
  val f1 = #$\overline{\text{fun}}$#((t: Rep[A], k1: Rep[B => C]) => #$\overline{\text{reset}}$#(#\underline{k1}#(#$\overline{\text{\Blue{f}}}$#(t))))
  #\underline{f1}#((y, #$\overline{\text{fun}}$#(k)))
}
\end{lstlisting}
\end{center}
With this @FUN@ subroutine, implementing a differentiable tree traversal is
straightforward.
We can define a @TREE@ abstraction to recursively traverse a @Rep[Tree]@ data structure.
For empty trees, the @init@ value is returned directly.
For non-empty trees, the function @b@ composes the recursive results from the subtrees.
\begin{lstlisting}[language=Scala,basicstyle=\footnotesize\ttfamily,escapechar=\#]
def #$\overline{\text{TREE}}$#(init: Rep[B])(t: Rep[Tree])(b: (Rep[B], Rep[B]) => Rep[B] @cps[Rep[C]]): Rep[B] @cps[Rep[C]] = {
  def #$\overline{\text{f}}$# = #$\overline{\text{FUN}}$# { tree: Rep[Tree] =>
    // If tree is not empty, recurse on subtrees and compose results, otherwise return the initial values.
    #$\overline{\text{IF}}$# (tree#\underline{.notEmpty}#) { #$\overline{\text{b}}$#(#$\overline{\text{f}}$#(tree#\underline{.left}#), #$\overline{\text{f}}$#(tree#\underline{.right}#)) } {init}
  }
  #$\overline{\text{f}}$#(t)
}
\end{lstlisting}

How do the above implementations relate to the formal $\texttt{letrec}$ rule?
The @FUN@ definition is similar to the representation of
$\lambda y. \ \orb\bra{\LetRec{f}{\lambda t. \ e} \At f y}$, which can be transformed as below:
\vspace{-5pt}
\begin{footnotesize}
$$
\ba{rll}
& &\SLam y. \ \orb\bra{\LetRec{f}{\lambda t. \ e} \At f y} \\
&=&\SLam y. \ \SShift{k}{\DLetRec{f}{\DLam t.\ \DLam k_1. \ \SRes{\Wat k_1 \ \orb\bra{e}}} \Dat (\Dat f \ y) \ (\WLam a. \ \Sat k \ a)} \\
&=&\SLam y. \ \SShift{k}{\DLetRec{f}{\DLam \hat t.\ \DLam k_1. \ \SRes{\Wat k_1 \ (\Sat \Blue{(\SLam t. \ \orb\bra{e})} \ \hat t)}} \Dat (\Dat f \ y) \ (\WLam a. \ \Sat k \ a)} \\
\ea
$$
\end{footnotesize}
\\[-5pt]
Note that in this term, we use \Blue{blue color} to highlight the sub-term that
corresponds to the parameter of @FUN@ (also marked in blue).
We can also see that the body of @TREE@ (i.e. $\overline{\texttt{f}}$(@t@) in the code)
evaluates to $\footnotesize{\orb\bra{\LetRec{f}{\lambda t. \ e} \At f\ t}}$.
Below is an example using the @TREE@ construct:
\begin{lstlisting}[language=Scala,basicstyle=\footnotesize\ttfamily]
def snippet(tree:Rep[Tree], in:Rep[Double]):Rep[Double] = grad(x => TREE(x)(tree){(l, r) => l * r * tree.value})(in)
\end{lstlisting}
The code after reducing @grad@, @FUN@, @TREE@, and some @shift@/@reset@ is shown below,
The generated pseudo LMS-IR and the generated C++ code are in Figure ~\ref{fig:treecode}.
\begin{lstlisting}[language=Scala,basicstyle=\footnotesize\ttfamily,escapechar=\#]
def snippet(tree: Rep[Tree], in: Rep[Double]): Rep[Double] = {
  val z = new NumR(in, 0.0)
  val k: Rep[NumR => Unit] = (x => x.d = 1.0)
  def #\underline{f1}#(t: Rep[Tree], k0: NumR => Unit) = #\underline{if}# (t.notEmpty) {
    val k_l: Rep[NumR => Unit] = (l =>
      // elide process of reset block (similar to straight-line program)
      val k_r: Rep[NumR => Unit] = (r => #$\overline{\text{reset}}$#{ #\underline{k0}#(l * r * t.value) })
      #\underline{f1}#(t.right, k_r))
    #\underline{f1}#(t.left, k_l)
  } #\underline{else}# k0(z)
  #\underline{f1}#(tree, k)
  z.d
}
\end{lstlisting}
\begin{figure}
\vspace{-1ex}
\begin{minipage}[t]{0.40\textwidth}
\begin{lstlisting}[language=Scala,basicstyle=\footnotesize\ttfamily]
def snippet(tree, in) = {
  d0 = ref 0
  k = (x, d) => d := 1.0
  f1 = (t, k0) => if (t.notEmpty) {
    k_l = (x_l, d_l) => {
      k_r = (x_r, d_r) => {
        v0 = t.value
        v1 = x_l * x_r * v0; d1 = ref 0
        k0(v1, d1)
        d_l += x_r * v0 * (! d1)
        d_r += x_l * v0 * (! d1)
      }
      f1(t.right, k_r)
    }
    f1(t.left, k_l)
  } else k0(in, d0)
  f1(tree, k)
  ! d0
}
\end{lstlisting}
\end{minipage}
\begin{minipage}[t]{0.50\textwidth}
\begin{lstlisting}[language=C,basicstyle=\footnotesize\ttfamily]
double Snippet(Tree tree, double in) {
  double d = 0.0;
  auto k = [&](double x, double& d) { d = 1.0; };
  auto rec = [&](Tree tree, function<void(double, double&)> k) {
    if (tree.notEmpty) {
      auto k_l = [&](double x_l, double& d_l) {
        auto k_r = [&](double x_r, double& d_r) {
          double x_t = tree.value; double dt = 0.0;
          k(x_l * x_r * x_t, dt);
          d_l += x_r * x_t * dt;
          d_r += x_l * x_t * dt;
        };
        rec(tree.right, k_r);
      };
      rec(tree.left, k_l);
    } else k(in, d);
  };
  rec(tree, k);
  return d;
}
\end{lstlisting}
\end{minipage}
\vspace{-3ex}
\caption{Generated pseudo LMS-IR (left) and the generated C++ code (right) of the tree example.}\label{fig:treecode}
\vspace{-5ex}
\end{figure}

With the above implementations, we have established a staged reverse-mode AD
framework that supports branches, loops, and recursion.
Though implementing these control-flow operators requires some engineering,
they simply combine CPS transformation with staging in the standard way (as shown in Figure~\ref{fig:reverse-formal2}).
The resulting framework provides a programming interface that is similar in style and
expressiveness to PyTorch.
It also generates an intermediate representation with inlined AD logic
(pure manipulation of @Doubles@ or @Tensors@)
which allows extensive optimizations similar in style to TensorFlow.

We note in passing that while it is, naturally, an option to implement CPS at the LMS IR level,
we choose to forgo this route in favor of the presented implementation for accessibility and simplicity.
A good, \emph{selective}, CPS transform (that transforms only the minimum necessary code to CPS)
is nontrivial to implement \cite{DBLP:conf/icfp/RompfMO09}.

\vspace{-1ex}
\section{Evaluation}\label{sec:eval}

So far, we have shown both how to implement plain PyTorch-style reverse-mode AD using delimited continuations,
and how to mix in multi-stage programming for TensorFlow-style graph reification.
Now, we extend our implementation to tensor operations,
and present
a system, Lantern\textsuperscript{\ref{github:lantern}}, that scales our described approach to real-world deep learning workloads.

The snippet below shows the basic structure of the staged tensor API:
\begin{lstlisting}[language=Scala,basicstyle=\footnotesize\ttfamily]
class Tensor(val data: Rep[Array[Double]], val dimension: Array[Rep[Int]]) {...}
class TensorR(val x: Tensor, val d: Tensor) {...}
\end{lstlisting}
The type of field @dimension@, @Array[Rep[Int]]@,
indicates that the tensor rank (number of dimensions) is always known at staging time.
The @TensorR@ class takes two @Tensor@s, one as the value, and the other as the gradient.
Operators in @TensorR@ are overloaded with @shift@ constructs, providing access 
to delimited continuations.
In analogy with the CPS-style implementation in Section~\ref{sec:reverse},
class @Tensor@ takes the role of @Double@, and @TensorR@ that of @NumR@.

The @Tensor@ class provides all tensor-level operations including element-wise operations with broadcasting,
matrix multiplication, convolution, and so on.
Lantern provides abstractions to run each of those operations either on CPU or GPU.
Implementations make use of BLAS library functions (for CPU) and cuBLAS/cuDNN library functions (for GPU),
but also include nested for loops (for CPU) and custom CUDA kernels (for GPU).

We would like to stress that the efficiency of Lantern can be further improved
by more sophisticated backend engineering, which is not the focus of this paper.
One direction is tensor IR level optimization similar to TVM
\citep{DBLP:conf/osdi/ChenMJZYSCWHCGK18} and Glow \citep{DBLP:journals/corr/abs-1805-00907},
including operator fusion, and systematic operation scheduling.
LMS provides fusion and array facilities 
\citep{DBLP:conf/popl/RompfSABJLJOO13} that have been used in OptiML \citep{DBLP:conf/icml/SujeethLBRCWAOO11}
and other DSLs based on the Delite compiler framework
\citep{DBLP:conf/IEEEpact/BrownSLRCOO11, DBLP:conf/cgo/BrownLRSSAO16}. These could be leveraged for Lantern as well.

Tensor IR level optimization is naturally supported by define-then-run systems (e.g. Lantern and TensorFlow),
but not by define-by-run systems (e.g. PyTorch, though recently PyTorch 1.0
moves towards this direction by extracting computation graphs using Torch Script \cite{TorchScript}).
Another important direction is advanced batching support, either in the form of \emph{autobatching} \`a la Dynet \citep{DBLP:conf/nips/NeubigGD17}
or \emph{dynamic batching} \`a la TensorFlow Fold \citep{DBLP:journals/corr/LooksHHN17}.
Advanced batching support is particularly useful in dynamic models where manual batching is challenging.
Another use is suggesting optimal batch sizes based on model and hardware details (e.g. GPU memory size).

Even with the current level of backend engineering, our evaluation shows that Lantern is competitive on contemporary
machine learning models, thus pushing the boundaries of existing frameworks in
various dimensions (expressivity and efficiency).

\subsection{Recursive Neural Network: TreeLSTM}\label{sec:tree}

In this and the following sections, we evaluate Lantern on several commonly used machine learning models.
Our evaluation focuses on expressivity (the ability to express various kinds of machine learning models)
and efficiency (the runtime for training those machine learning models on a single
GeForce GTX 1080 Ti with CUDA 10, using PyTorch version 1.0rc, TensorFlow version 1.12.0-rc0).
Runtime results are reported as the median runtime of 5 epochs with 3 repeats.
The correctness of the computed gradients is implicit (we have extensive unit tests, and we check our gradients with PyTorch).
We elide loss curves and other dimensions of evaluation such as hyperparameter-tuning/cross-validation/testing.

We start the evaluation with TreeLSTM, which is a state-of-the-art \emph{recursive} machine learning model that
heavily depends on dynamic control flow guided by structural training data.
Models like this are useful for handling natural language parse trees and abstract syntax trees of programming languages.
At the same time, such dynamic models pose interesting challenges to machine learning frameworks.

We showcase TreeLSTM on the Sentiment Classification task \citep{DBLP:journals/corr/TaiSM15}
using the Stanford Sentiment Treebank dataset \cite{treebank}.
The dataset contains sentences of movie reviews, which are parsed into binary trees based on language semantics.
Each leaf node contains a word that can be mapped to a known numeric vector (embedding) based on word semantics.
TreeLSTM should model a function (@Bi-LSTM@) that recursively computes hidden states for all nodes,
and a mapping from hidden states to sentiment scores, both of which are then
trained end-to-end by minimizing \emph{softmax-cross-entropy loss} with regard to true sentiment labels of each node.

The recursive function that computes hidden states of node $i$ can be written as below:
$$\ h_i = \text{Bi-LSTM}(\text{Embedding}(i.\text{word}), h_{i.\text{left}}, h_{i.\text{right}})$$
where $h_i$ represents hidden state of node $i$, @Embedding@ represents the known mapping from words to their embeddings,
$i.\text{word}, \ i.\text{left}, \ i.\text{right}$ represent the optional word, left-child, and right-child of node $i$,
and @Bi-LSTM@ is a variant of LSTM that can handle two hidden states as inputs.

It is easy to express this model in Lantern since Lantern supports unrestricted control flow including branches, loops, and recursion.
We show the core of the TreeLSTM model in Lantern (function @lossFun@) below.
The @TREE@ abstraction (Section~\ref{sec:tree_abs}) makes the 
code very concise. Users merely need to supply
the (anonymous) function that computes the state of the current node from the states of the left and right children (@lState@/@rState@):

\begin{lstlisting}[language=Scala,basicstyle=\footnotesize\ttfamily]
def lossFun(node: Rep[Tree]) = {
  val initialState = State(loss = 0, hidden = 0, cell = 0)
  val resultState  = TREE(initialState)(node) { (lState, rState) =>
    val embedding      = IF (node.isLeaf) { Embedding(node.word) } { 0 }
    val (hidden, cell) = BiLSTM(embedding, lState, rState)
    val loss           = softmaxCrossEntropyLoss(Linear(hidden), node.score)
    State(loss, hidden, cell)
  }
  resultState.loss
}
\end{lstlisting}

\begin{figure}[htpb!]
\vspace{-2ex}
\begin{minipage}[t]{0.50\textwidth}
  \includegraphics[width=\linewidth,scale=1.0]{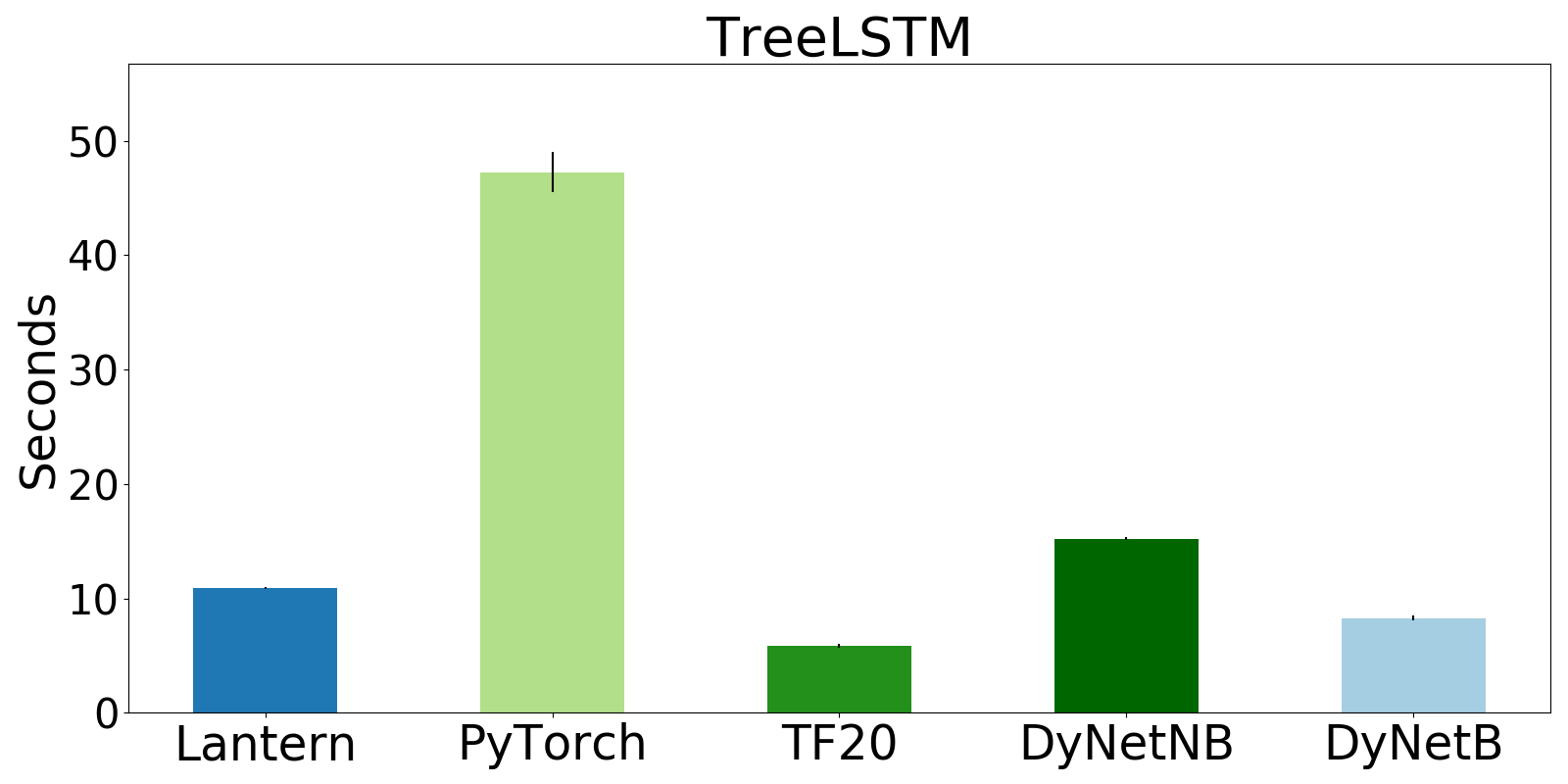}
\end{minipage}
\vspace{-3ex}
\caption{Running time of TreeLSTM for different frameworks.}
\label{fig:TreeLSTM}
\vspace{-3ex}
\end{figure}

This model can also be expressed easily in define-by-run frameworks like PyTorch where computation graphs are constructed on-the-fly (via Python recursive functions in this case).
However, as we can see from the runtime (Figure~\ref{fig:TreeLSTM}),
training TreeLSTM in PyTorch is very slow (more than 4 times slower than Lantern),
mainly due to the overhead incurred for each individual
computation step. In comparison, another define-by-run framework called DyNet \citep{DBLP:conf/nips/NeubigGD17}
has a more lightweight internal graph representation and optimized C++ backend \citep{DBLP:journals/corr/NeubigDGMAABCCC17},
which makes it run faster than PyTorch (see DyNetNB, short for DyNet-No-Batching),
though still about 1.6 times slower than Lantern.

On the other hand, TensorFlow has trouble expressing TreeLSTM, or any other recursive neural network models.
This is mainly due to limitations of TensorFlow's static graph construction interface,
which does not support recursion. As a consequence, TensorFlow can neither define static
computation graphs that are recursive and covers structural data of different shapes (like Lantern),
nor define computation graphs dynamically based on each structural data (like PyTorch).
Other ways to flatten structural data into sequences and model them with recurrent neural networks
often incur high memory overhead.
Interestingly, TensorFlow Fold \citep{DBLP:journals/corr/LooksHHN17},
a library on top of TensorFlow, manages to push this limited static computation graph even further for training
recursive neural networks such as TreeLSTM. The main idea is that, given a set of static computation graphs
of different shapes, TensorFlow Fold rewrites them into one static computation graph that handles all
given graphs, by extensively using extra @concat@ and @gather@ operations to move data around.
The added benefit is that instances of the same operations at the same depth can be batched together
(in machine learning, batching refers to processing multiple pieces of data simultaneously, often in a mini-batch),
which makes training more efficient. Indeed TensorFlow Fold (TF20, short for TensorFlow Fold at batch size 20)
is the most efficient framework in our evaluation of TreeLSTM.

The dynamic batching approach of TensorFlow Fold is not the only way to batch training data for recursive neural networks.
DyNet provides another strategy called autobatching. Being a dynamic framework,
DyNet has the freedom to construct and manipulate computation graphs on-the-fly, including automatically batching
nodes in computation graphs based on node types, data dimensions, and node dependencies. However, it should be
noted that using autobatching efficiently requires good batching heuristics from the framework, and some input
from the user, who controls the partitioning of computation graphs considered for autobatching. In our case,
DyNetB (short for DyNet-Batching) shows about 50\% improvement on GPU runtime when we allow autobatching
within each input structure, but not across multiple input structures.
Although both batched frameworks outperform Lantern, it should be noted that dynamic batching
(as in TensorFlow Fold) could also be added to Lantern with additional engineering effort.

\subsection{Convolutional Neural Networks: SqueezeNet and ResNet50}\label{sec:cnn}

For non-recursive models, batching simply means adding an extra dimension to the input data,
which is supported in Lantern. We now evaluate representative convolutional neural networks.

SqueezeNet \citep{DBLP:journals/corr/IandolaMAHDK16} and ResNet50 \citep{DBLP:conf/cvpr/HeZRS16}
are contemporary convolutional neural network models for image classification.
SqueezeNet uses a carefully designed CNN architecture so that it contains fewer parameters, but
shows a similar level of accuracy as larger models.
ResNet50 belongs to the ResNet family of CNN architectures, which makes use of batch-normalization,
residual connections, and other techniques for fast/stabilized training.

\begin{figure}[h!]
\vspace{-2ex}
\begin{center}
\begin{minipage}[t]{0.35\textwidth}
  \includegraphics[width=\linewidth,scale=1.0]{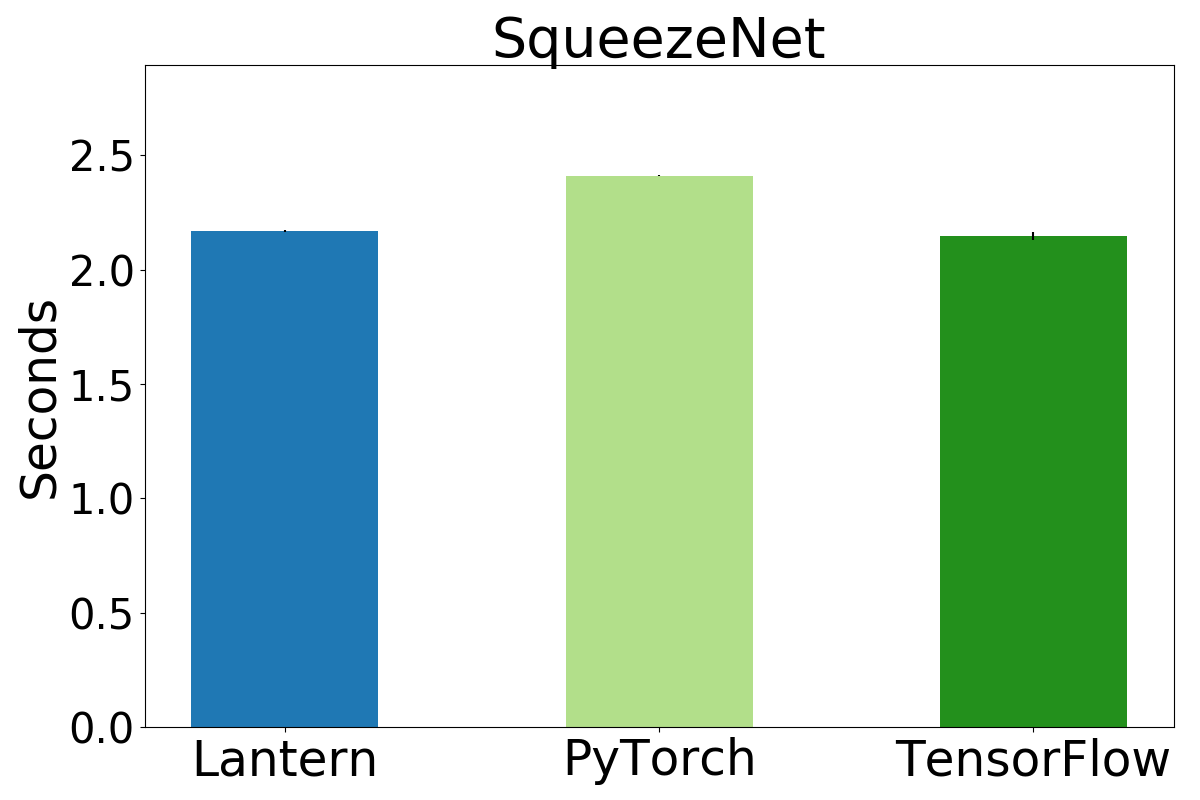}
\end{minipage}%
\begin{minipage}[t]{0.35\textwidth}
  \includegraphics[width=\linewidth,scale=1.0]{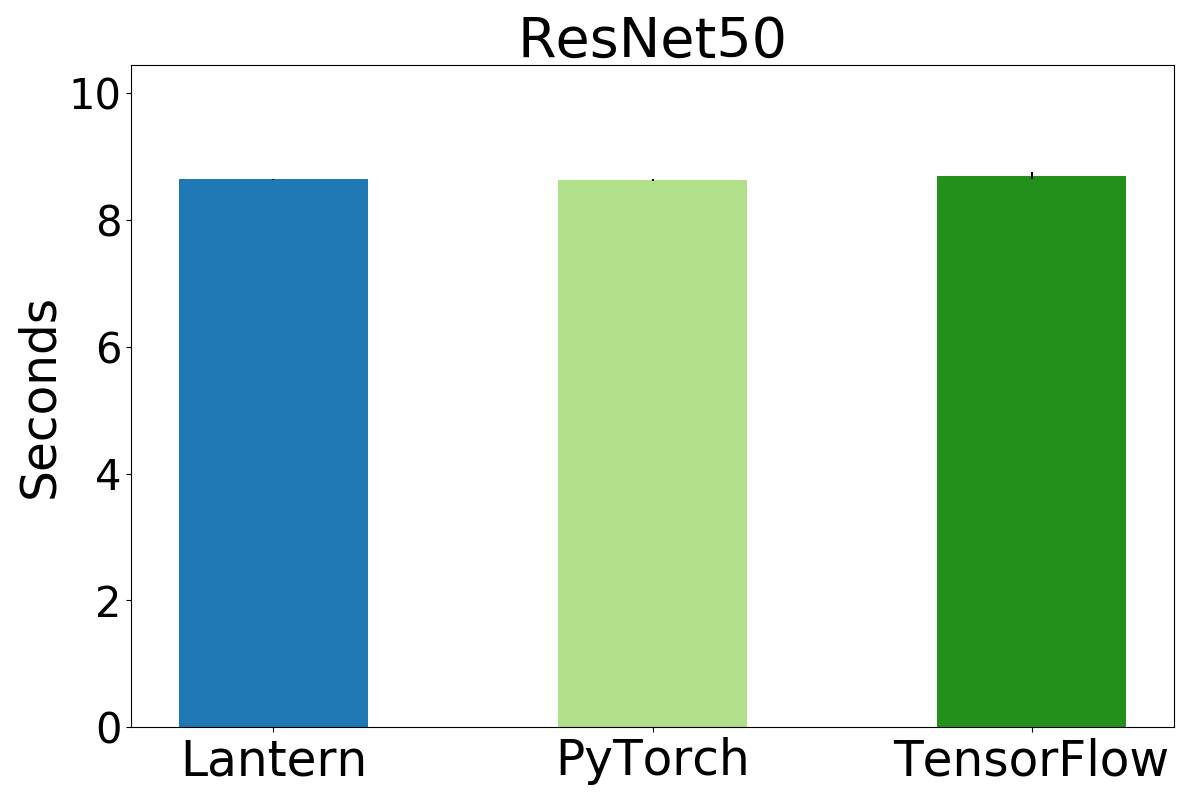}
\end{minipage}
\end{center}
\vspace{-3ex}
\caption{Running time of SqueezeNet and ResNet50 for different frameworks.}
\label{fig:CNNEval}
\vspace{-3ex}
\end{figure}

PyTorch and TensorFlow implementations of these models exist on GitHub.
We hand-wrote identical Lantern models and also imported existing ONNX (Open Neural Network Exchange) models into Lantern.
We evaluate these implementations on the CIFAR-10 dataset \citep{cifar10}.
As shown in Figure~\ref{fig:CNNEval},
Lantern and TensorFlow perform slightly better than PyTorch on SqueezeNet,
and all three models have similar runtime performance on ResNet50.
This result is expected since SqueezeNet and ResNet50 are mostly composed of convolution layers,
which dominate the runtime cost.
Convolution layers heavily rely on hardware-specific library functions such as
cuDNN API functions, rendering other graph-level optimizations (of TensorFlow and potentially Lantern) insignificant.
However, getting Lantern on par with PyTorch and TensorFlow took
non-trivial effort: memory management techniques were crucial when using cuDNN API functions.

\vspace{-2ex}
\subsection{DeepSpeech2}\label{sec:rnn}

DeepSpeech2 \cite{DBLP:journals/corr/AmodeiABCCCCCCD15} is a representative deep
neural network for automatic speech recognition (ASR), which reaches
state-of-the-art performance on real-world datasets. DeepSpeech2 is the most
complex model in our evaluation: it is a real production model with
convolutional, batch norm, and RNN layers, and is trained with the CTC
(Connectionist Temporal Classification) loss function. A variant
of this model is included in the MLPerf benchmark suite \citep{MLPerf}. 

\begin{wrapfigure}{r}{0.23\textwidth}
  \vspace{-3ex}
  \begin{center}
    \includegraphics[width=0.23\textwidth]{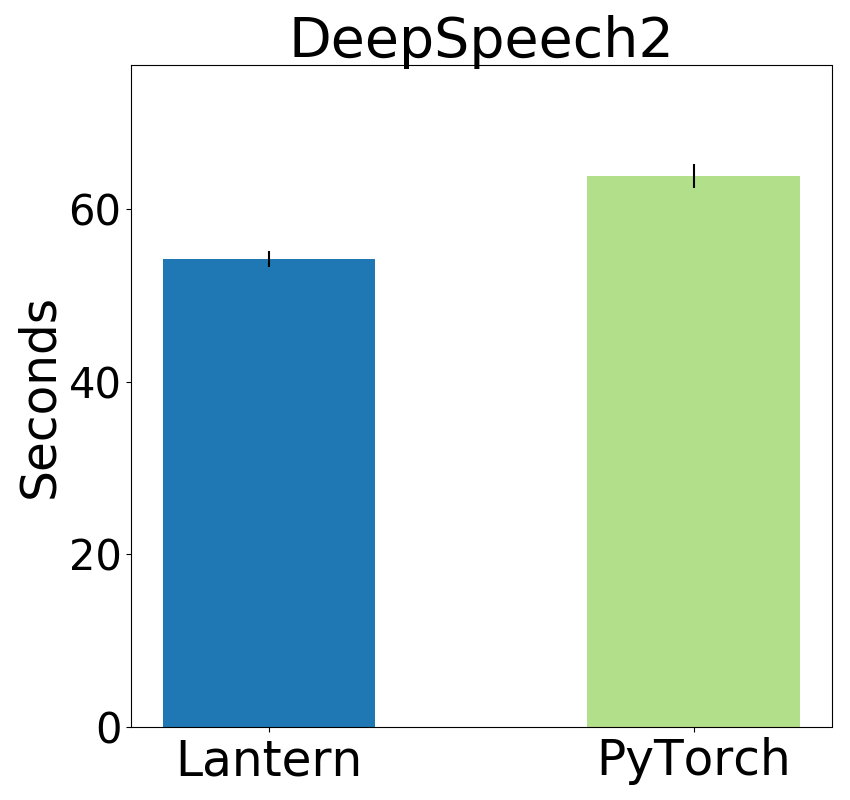}
  \end{center}
  \vspace{-3ex}
  \caption{Running time of DeepSpeech2 for different frameworks.}
  \label{fig:DeepSpeech2}
  \vspace{-3ex}
\end{wrapfigure}

We evaluated DeepSpeech2 models on the Librispeech \cite{Panayotov2015LibrispeechAA} dataset,
but skipped TensorFlow because it uses a custom CPU implementation of @CTCLoss@, making a fair comparison impossible.
Lantern and PyTorch models both use bidirectional RNNs with ReLU activation and SGD with momentum.

At the time of writing, Lantern is $\sim$10\% faster than PyTorch (Figure~\ref{fig:DeepSpeech2}) on this model,
most likely because Lantern spent extra time to select better CuDNN kernel functions for CNN in the first mini-batch.
Additional engineering can probably further improve Lantern's efficiency, via better hand-written kernel functions
(such as softmax, activation, and so on). However, surpassing the performance of existing frameworks on these
well-tuned model implementations has not been our primary goal; the aim of this paper was merely to demonstrate the scope of Lantern,
and to validate our fundamental design.

\section{Related Work}\label{sec:related}

\paragraph{Automatic Differentiation: A History}
Gradient-based optimization lies at the heart of machine learning, with backpropagation~\cite{rumelhart1986learning},
an application of differentiation, as a key ingredient for training neural networks.
The fundamental idea of automatic differentiation (AD) emerged in the 1950s as programs that calculate
derivatives alongside the normal computation~\cite{nolan1953thesis, Beda1959Pfa}. A formal introduction to forward-mode AD appeared in the
1960s~\cite{DBLP:journals/cacm/Wengert64}.
The application of gradient descent to large-scale optimization first arose in control theory \cite{bryson1962steepest, bryson1975applied}, although the underlying ideas are of course much older. In the
1970s, \citet{linnainmaa1976taylor} introduced the concept of reverse-mode AD and the related idea of computation
graphs, which are now widely used by modern machine learning frameworks. \citet{speelpenning1980compiling} implemented
reverse-mode AD in a general-purpose language, which is considered the first implementation of reverse-mode
AD that performed gradient computations automatically. At the same time, backpropagation was invented and reinvented
within the machine learning community~\cite{werbos1974beyond, parker1985learning, rumelhart1986learning}. This
divergence continued until \citet{DBLP:journals/nn/Hecht-Nielsen88a} brought together the work from both communities.

\paragraph{Automatic Differentiation: A PL View}
AD has also received attention from the programming language community, with recent proposals to generalize neural
network models to differentiable functional programs~\cite{blog:nn_fp, fong2017backprop}. This development is also
fueled by modern deep learning frameworks, which define neural networks ``very much like a regular program''
\cite{blog:differentiable_programming, tensorflowsemantics}. Some recent research demonstrates this direct correspondence between the two
fields by implementing differentiable analogs of traditional data structures~\cite{NIPS2015_5648} and machine
models \cite{DBLP:journals/corr/GravesWD14}.
Another line of work has aimed to formalize AD, both forward-mode \cite{DBLP:journals/lisp/SiskindP08} and reverse-mode
\cite{DBLP:journals/toplas/PearlmutterS08}. There exist high-level languages with first-class AD operators
\cite{DBLP:journals/corr/SiskindP16a}, as well as flexible AD library implementations, e.g., DiffSharp
\cite{DBLP:journals/corr/BaydinPS16}. A Haskell implementation of forward-mode AD was proposed by
\citet{Elliott2009-beautiful-differentiation}. Swift for TensorFlow \cite{SwiftForTensorFlow}
integrates AD as a first-class feature in a general purpose language.
\citet{DBLP:journals/corr/abs-1806-02136} demonstrated that forward-mode AD can sometimes outperform reverse-mode AD
when combined with aggressive fusion and code motion techniques in a functional array programming language.
This could also be achieved in Lantern using existing fusion and array facilities in LMS
\citep{DBLP:conf/popl/RompfSABJLJOO13}, as used in OptiML \citep{DBLP:conf/icml/SujeethLBRCWAOO11}
and other DSLs based on the Delite compiler framework
\citep{DBLP:journals/corr/abs-1109-0778, DBLP:conf/IEEEpact/BrownSLRCOO11, DBLP:conf/cgo/BrownLRSSAO16}.
\citet{DBLP:journals/corr/BaydinPR18} provided a thorough review of AD and deep learning from a functional programming perspective.

\paragraph{A Tale of Two Styles}
Most modern deep learning frameworks compute gradients of training loss with respect to neural network parameters
in one of two ways \cite{DBLP:journals/corr/BaydinPR18}.
The first is to let users define computation graphs using a domain-specific language (DSL) and to interpret graph
operations at runtime. Computation graphs represent entire programs and are more amenable to global analysis and
optimizations like operator fusion. However, graph-building DSLs are limited in expressivity, contain unintuitive
control structures, and are difficult to debug. Frameworks such as Theano~\cite{DBLP:journals/corr/Al-RfouAAa16} and
TensorFlow~\cite{DBLP:journals/corr/AbadiABBCCCDDDG16} belong to this category. The other way is to integrate
general-purpose programming languages with reverse-mode AD as a library, of which Torch~\cite{Collobert_NIPSWORKSHOP_2011},
PyTorch~\cite{paszke2017pytorch, paszke2017automatic}, Autograd~\cite{maclaurin2016modeling}, and Chainer~\cite{tokui2015chainer}
are well-known representatives. Caffe~\cite{DBLP:journals/corr/JiaSDKLGGD14}, MXNet~\cite{chen2015mxnet}, and CNTK~\cite{seide2016cntk}
are somewhere in the middle. The tight integration between host languages and AD frameworks of the pure-library
category has certain usability benefits, such as natural control flow and easy debugging, but comes at the expense of efficiency.
Neural network exchange formats such as ONNX~\cite{onnx2017} aim to bridge this
gap by enabling an easy conversion between frameworks.

Previous attempts at building source-to-source deep learning compilers mostly focus on either the define-by-run or
define-then-run approach, as noted by \citet{DBLP:journals/corr/BaydinPR18}. Tangent \cite{tangent2017, 2017arXiv171102712V}
implements a source-to-source compiler in Python which supports automatic differentiation, but this framework
constrains the host language to a limited subset of Python. DLVM
\cite{dlvm2017, DBLP:journals/corr/abs-1711-03016} compiles deep learning programs
written in Swift into a domain-specific SSA IR, performs analyses and transformations
(including source code transformation AD), and generates code via LLVM.
Swift for TensorFlow \cite{SwiftForTensorFlow} mixes the two approaches:
it enables imperative-style programs but uses a ``graph program extraction''
compiler transform to automatically extract tensor code and build computation
graphs.

\paragraph{Staging: A Unification of the Two Styles}
The present work aims to reap the benefits of both styles by using a computation graph DSL that really is a 
general-purpose programming language. Our transformation of high-level neural networks to low-level code is fueled by the idea of
multi-stage programming (staging). More than 30 years ago, \citet{DBLP:conf/popl/JorringS86} observed that
many computations can be naturally separated into stages distinguished by frequency of execution or availability of
data. The idea to treat staging as an explicit \emph{programming model} was popularized, among others, by
\citet{DBLP:journals/tcs/TahaS00}. Since then, modern staging approaches blend normal program
execution with the delayed construction of an \emph{intermediate program representation} (IR), which may be a computation
graph, or in  more traditional systems, an abstract syntax tree (AST). We use the Lightweight Modular Staging (LMS) framework
\citep{DBLP:conf/gpce/RompfO10}, which provides a rather seamless implementation of staging in the Scala language and
has been utilized in a range of existing applications~\cite{DBLP:conf/icml/SujeethLBRCWAOO11, DBLP:conf/snapl/RompfBLSJAOSKDK15, DBLP:conf/icfp/RompfA15}.

\paragraph{Delimited Continuations: A Simpler Essence}
Lantern relies on delimited continuations~\cite{DBLP:journals/mscs/DanvyF92, DBLP:journals/tcs/DanvyN03, DBLP:conf/lfp/DanvyF90},
as implemented in Scala~\cite{DBLP:conf/icfp/RompfMO09}.
In parallel to our work, which first appeared as tech report on arXiv~\cite{DBLP:journals/corr/abs-1803-10228},
\citet{DBLP:journals/pacmpl/Elliott18} proposed a generalized view of AD based on the paradigm of
``compiling to categories'' \cite{DBLP:journals/pacmpl/Elliott17a}.
The paper echoes our view of AD as a specific form of symbolic differentiation and also
mentions continuations for reverse AD, but overall it approaches the problem from a very different categorical
perspective.
In comparison, our work proposes what we think is an ``even simpler essence''
of automatic differentiation. In particular, we show that continuations are central
to reverse-mode AD, but that category theory is optional. Focusing on continuations
as the key enabler makes reverse-mode AD (and hence gradient-descent optimization)
immediately applicable to basically any program, including in-graph recursion and
higher-order functions.

\section{Conclusions}\label{sec:conclusions}

With this paper, we set out to demystify automatic differentiation by examining 
it through the lens of program transformation. We established a tight connection
between reverse-mode AD and delimited continuations.
With the help of delimited continuation control operators,
we provided an implementation of reverse-mode AD using operator overloading
that is no more complex than forward-mode AD.

We further combined this formulation of AD with multi-stage programming (staging), 
which leads to a highly efficient implementation that combines the performance 
benefits of deep learning frameworks based on explicit reified computation graphs 
(e.g., TensorFlow) with the expressivity of pure library approaches (e.g., PyTorch).

Based on these two ideas, we have built a deep learning framework named
Lantern. With native C++/CUDA backends, Lantern attains competitive performance
for a variety of state-of-the-art deep learning models,
such as SqueezeNet, ResNet, DeepSpeech2, and TreeLSTM.

\section*{Acknowledgements}\label{sec:ack}

We thank the anonymous reviewers and especially our anonymous shepherd for numerous
thorough and thoughtful comments and suggestions, especially the suggestion of encoding
multi-level CPS via nesting explicit CPS transformation in @shift/reset@.
This work was supported in part by NSF awards 1553471 and 1564207, DOE award DE-SC0018050,
as well as gifts from Google, Facebook, and VMware.

\clearpage
{
\bibliography{references}
}

\end{document}